\documentclass[letterpaper]{article} 
\usepackage{aaai23}  
\usepackage{times}  
\usepackage{helvet}  
\usepackage{courier}  
\usepackage[hyphens]{url}  
\usepackage{graphicx} 
\usepackage{natbib}  
\usepackage{caption} 
\usepackage{algorithm}

\usepackage{newfloat}
\usepackage{listings}
\DeclareCaptionStyle{ruled}{labelfont=normalfont,labelsep=colon,strut=off} 
\floatstyle{ruled}
\newfloat{listing}{tb}{lst}{}
\floatname{listing}{Listing}
\title{Multi-Level Compositional Reasoning for Interactive Instruction Following \\
    \vspace{0.3em}
    \vspace{-0.5em}
}
\author{
    Suvaansh Bhambri\equalcontrib,  
    Byeonghwi Kim\equalcontrib,
    Jonghyun Choi 
}
\affiliations{
    Yonsei University\\

    suvaanshbhambri@gmail.com, 
    byeonghwikim@yonsei.ac.kr,
    jc@yonsei.ac.kr
}

\usepackage{bibentry}

\usepackage{url}

\usepackage[cal=pxtx,frak=pxtx]{mathalfa}
\usepackage{arydshln}
\usepackage{booktabs}
\usepackage{enumitem}
\usepackage{balance}
\usepackage{combelow}
\usepackage{tabularx}

\usepackage{wrapfig}
\usepackage[stable]{footmisc}

\usepackage{multirow}
\usepackage{makecell}
\usepackage{mathtools}
\usepackage{graphicx}
\usepackage{algorithmicx}
\usepackage{eqparbox,array}

\usepackage{xcolor,colortbl}
\usepackage{subcaption}
\usepackage{caption}

\usepackage{mwe} 
\usepackage{calc} 
\usepackage{pifont} 

\newcommand{\method}{\mbox{{MCR-Agent}}\xspace}
\newcommand{\methodfull}{{Multi-level Compositional Reasoning Agent}\xspace}

\newcommand{\cmark}{\color{black}{\ding{51}}}
\newcommand{\xmark}{\color{black}{\ding{55}}}

\makeatletter
\DeclareRobustCommand\onedot{\futurelet\@let@token\@onedot}
\def\@onedot{\ifx\@let@token.\else.\null\fi\xspace}

\def\eg{\emph{e.g}\onedot} 
\def\ie{\emph{i.e}\onedot} 
 
\def\etc{\emph{etc}\onedot} \def\vs{\emph{vs}\onedot}

\makeatother

\newcommand{\rev}[1]{\textcolor{black}{#1}} 
\newcommand{\revv}[1]{\textcolor{black}{#1}} 
\newcommand{\reva}[1]{\textcolor{black}{#1}} 

\begin{document}

\maketitle

\begin{abstract}
    Robotic agents performing domestic chores by natural language directives are required to master the complex job of navigating environment and interacting with objects in the environments.
    The tasks given to the agents are often composite thus are challenging as completing them require to reason about multiple subtasks, \eg, bring a cup of coffee.
    To address the challenge, we propose to divide and conquer it by breaking the task into multiple subgoals and attend to them individually for better navigation and interaction.
    We call it \emph{\methodfull (\method)}.
    Specifically, we learn a three-level action policy.
    At the highest level, we infer a sequence of human-interpretable subgoals to be executed based on language instructions by a high-level \emph{policy composition controller}.
    At the middle level, we discriminatively control the agent's navigation by a \emph{master policy} by alternating between a navigation policy and various independent interaction policies.
    Finally, at the lowest level, we infer manipulation actions with the corresponding object masks using the appropriate \emph{interaction policy}.
    Our approach not only generates human interpretable subgoals but also achieves 2.03\% absolute gain to comparable state of the arts in the efficiency metric (PLWSR in unseen set) without using rule-based planning or a semantic spatial memory.
\end{abstract}

\section{Introduction}
For the long-awaited dream of building a robot to assist humans in daily life, we now witness rapid advances in various embodied AI tasks such as visual navigation~\cite{anderson2018vision,chen2019touchdown,krantz2020navgraph}, 
object interaction \cite{zhu2017visual,misra2017mapping},
and interactive reasoning \cite{embodiedqa,gordon2018iqa}.
Towards building an ideal robotic assistant, the agent should be capable of all of these tasks to address more complex problems.
A typical approach for combining these abilities is to build a unified model \cite{shridhar2020alfred,singh2021factorizing} to jointly perform different sub-tasks. 
However, the reasoning for navigation can differ significantly from the \reva{one} for object interaction; the former needs to detect navigable space and explore to reach a target location while the latter requires detecting objects and analysing their distances and states~\cite{singh2021factorizing}.

Meanwhile, the human cognition process learns to divide a task into sub-objectives such as navigation or interaction, which enables humans to facilitate complex reasoning in various circumstances
\cite{HayesRoth1979ACM}.
Inspired by this, we propose a multi-level compositional reasoning agent (\method) that disentangles the task into high-level subgoals; then learns and infers a low-level policy for each sub-task.
Specifically, we propose a multi-level agent comprised of (1) a policy composition controller (PCC) that specifies a sub-policy sequence, (2) a master policy (MP) that specialises in navigation, and (3) a set of interaction policies (IP) that execute interactions.
This disentanglement enables easier analysis of subtasks with shorter horizons (See Sec. `Multi-Level Policy vs. Flat Policy' for empirical evidence). 

\reva{
In addition, to interact with multiple objects in a long sequence, the agent should be able to keep track of the current target object at each time instance.
Inspired by \cite{yang2018visual,wortsman2019learning}, we additionally propose an object encoding module (OEM) that provides target object information which is used as a navigational subgoal monitor, \ie, stopping criterion for the navigation policy. 
}

In our empirical evaluations with a long horizon instruction following task with the condition of not requiring additional \reva{depth} supervision and \reva{perfect egomotion} assumption, usually not available for real world deployment, we observe that \method outperforms most prior arts in literature by large margins.
We summarize our contributions as follows:

\begin{figure*}[t]
    \centering
    \includegraphics[width=.7\linewidth]{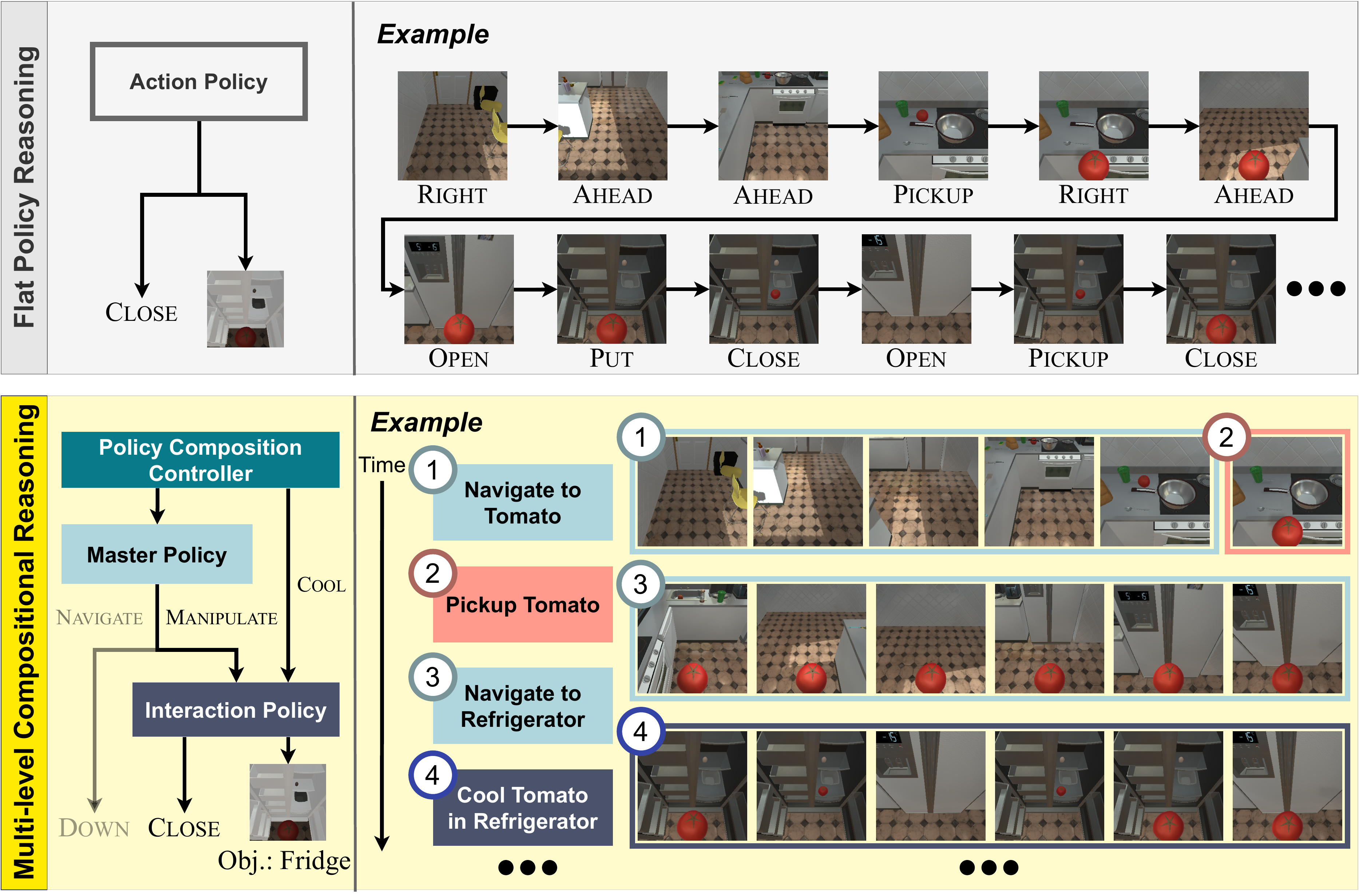}
    \vspace{-0.5em}
    \caption{
        The proposed `Multi-level compositional \reva{reasoning}' contrasted to `Flat policy \reva{reasoning}'. 
        The flat policy \reva{reasoning} has been employed in prior arts~\cite{shridhar2020alfred,singh2021factorizing,pashevich2021episodic,nguyen2021look}, training an agent to directly learn the low-level actions. 
        On the contrary, our multi-level policy decomposes a long-horizon task into multiple subtasks and leverages the high-level abstract planning, which enables an agent to better address long-horizon planning.}
    \label{fig:teaser}
    \vspace{-0.7em}
\end{figure*}

\begin{itemize}[leftmargin=10pt]
    \item[$\bullet$] We propose a multi-level hierarchical framework, \method, that decomposes a compositional task into semantic subgoals and effectively addresses them with corresponding submodules.
    \item[$\bullet$] We propose an object encoding module (OEM) that encodes object information from natural language instructions for effective navigation.
    \item[$\bullet$]
    By extensive quantitative analyses on a challenging interactive instruction following benchmark \cite{shridhar2020alfred}, we show that \method yields competitive performance with higher efficiency than prior arts that do not assume perfect egomotion and extra depth supervision.
\end{itemize}

\section{Related Work}
\label{sec:related_works}
There are numerous task setups and benchmarks proposed for developing an agent to complete complicated tasks given natural language directives, such as agents trained to navigate \cite{li2020mapping,xu2021grounding} or solve household tasks \cite{shridhar2020alfred}. 
However, the vast majority of approaches for these tasks employ flat reasoning \cite{singh2021factorizing,nguyen2021look}, in which the agent decides on the low-level actions accessible while moving through the environment \cite{gupta2017cognitive,zhu2020vision}. 
When the prior arts seek to define subtasks, some define them with two layers of hierarchy \cite{zhang2021hierarchical,corona2020modular,blukis2021persistent,andreas_modular_2017,gordon2018iqa,yu2019multi,das_neural_2019}. 
However, these strategies require a good amount of data due to the semantic gap between abstract natural language instructions and concrete executions \cite{zhou2021hierarchical}. Natural language is subjective and even a seemingly simple command can contain several unstated meanings. 
Because of this semantic gap, most approaches~\cite{landi2019embodied,krantz2020navgraph,singh2021factorizing,pashevich2021episodic} require either a large amount of labeled data or trial-and-error learning to map language to low-level actions.
In contrast, we propose to use deeper hierarchical knowledge for better control of embodied agents.
Thanks to the modular structure, our agent reasons and accomplishes tasks along longer paths, spanning numerous subgoals.

The described task requires not only navigation but also interaction.
\cite{shridhar2020alfred} proposes a CNN-LSTM-based baseline agent with progress tracking \cite{ma2019selfmonitoring}.
\cite{singh2021factorizing} offers a modular strategy  for factorising action prediction and mask generation while  \cite{nottingham2021modular} offers a system that encodes language and visual state, and performs action prediction using independently trained modules.
\cite{zhang2021hierarchical} propose a transformer-based hierarchical agent whereas \cite{suglia2021embodied} presents a transformer-based agent that uses object landmarks for navigation. \cite{pashevich2021episodic} also presents a transformer-based agent that uses a multimodal transformer for exploiting the multiple input modalities.

Recent work proposes to construct semantic maps and leverage the relative localization for improved navigation where
\cite{blukis2021persistent} uses a 3D map to encode spatial semantic representation, \cite{min2021film} suggests a SLAM-based approach that keeps observed information in a 2D top-down map while \cite{liu2022planning} presents a planning-based approach that keeps a semantic spatial graph to encode visual inputs and the agent's poses.

Finally, a modular policy with two levels of hierarchy has been proposed by \cite{corona2020modular} which does not perform well on a long-horizon task.
In contrast, our policy operates at three hierarchical levels, exploiting the fact that navigation and interaction are semantically diverse activities that require independent processing.

\section{Model}
\label{sec:hierarchical_policy}

\begin{figure*}[t]
    \centering
    \includegraphics[width=\linewidth]{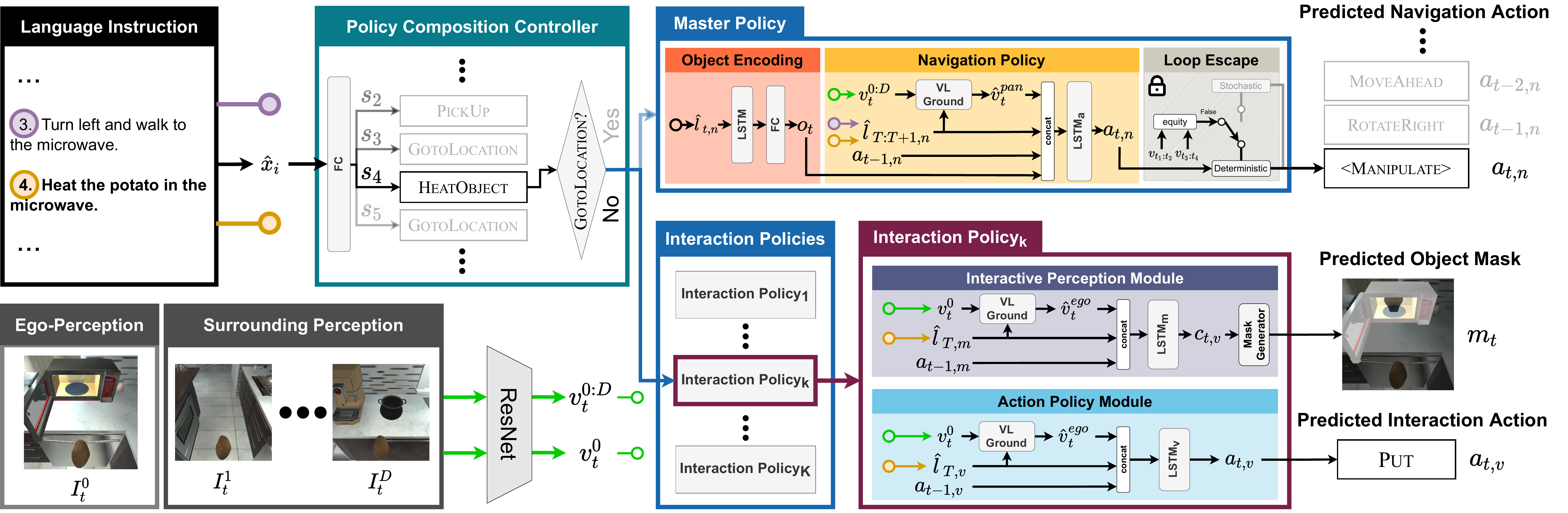}
    \vspace{-1.5em}
    \caption{
        Model Architecture.
        $I_t^d$ denotes an RGB frame from an explorable direction, $d \in [0, D]$, at the time step, $t$, where $d = 0$ indicates the egocentric direction.
        We encode $I_t^d$ using a pretrained ResNet and acquire a visual feature, $v_t^d$.
        $\hat{x}_i$ denotes each step-by-step instruction.
        $\hat{l}_{T,v}$, $\hat{l}_{T,m}$ denotes the encoded instruction for the `interactive perception module' and `action prediction module' respectively. $\hat{l}_{T:T+1,n}$ denotes the encoded `subtask' instruction (Sec. `Master Policy').
        $T$ refers to the index of the current subgoal.
        In our master policy, OEM outputs object encoding, $o_t$, using $\hat{l}_{T:T+1,n}$.
        `VL-Ground' uses dynamic filters to capture the correspondence between visual and language features and outputs attended visual features, $\hat{v}_{t}^{pan}$ and $\hat{v}_{t}^{ego}$.
    }
    \label{fig:overview}
    \vspace{-1em}
\end{figure*}

Observing that the visual information for navigation considerably varies over time while interacting with objects is largely stationary, we argue that the agent benefits from learning different policy modules for these two different tasks as follows.
The navigation needs to reason about the temporal history and global environment information. 
The interaction with objects requires focusing on local visual cues for precise object localization.
In addition, there is a sample imbalance between navigation and interaction actions as navigation actions are far more frequent than interaction actions.
This would bias a learned model towards more frequent actions, \ie, navigation.

Based on these observations, we design an architecture with three levels of compositional learning; (1) a high-level policy composition controller (PCC) that uses language instructions to generate a sequence of sub-objectives, (2) a master policy that specialises in navigation and determines when and where the agent is required to perform interaction tasks, and (3) interaction policies (IP) that are a collection of subgoal policies that specialise in precise interaction tasks.

Specifically, \method first analyzes each language instruction and uses the information to determine the basic high-level policy sequence required to perform the task. 
Following the predicted sequence, the control of the agent is shifted between (1) the master policy and (2) different interaction policies for object interaction.
Moreover, all interaction policies are compositional and independent, which allows formulating an instance-specific high-level action sequence. 
In particular, we learn multiple interaction policies, each of which specialises in a different sub-objective and can be integrated in a precise order to complete long-horizon tasks.
We illustrate the model overview in Fig. \ref{fig:overview}.

\subsection{Policy Composition Controller}
\label{sec:PCC}
The nature of the long-horizon instruction following is highly complex.
To address this, we argue that it is beneficial to first generate a high-level subgoal sequence, and then tackle each subgoal individually. 
Specifically, the trajectories are first divided into meaningful subgoals based on the given language instruction (called `step-by-step' instruction)~\cite{shridhar2020alfred}.
For inferring the subgoals, we propose a policy composition controller (PCC), shown as dark cyan box in Fig. \ref{fig:overview}, that predicts a subgoal $\mathcal{S} = \{s_i\}$ (where $s_i$ belongs to a set of predefined subgoals)
for each `step-by-step' instruction.
The PCC's predictions correlate to semantic subgoals, subjecting the agent's logic to observation. It gives the intuition on what the agent is attempting to accomplish at any particular instance. This enables us to track the progress of task completion by the agent.

Specifically, we first encode the language instructions with a Bi-LSTM, followed by a self-attention module.
Each encoded step-by-step language instruction $\hat{x}_i$ is used as input for the PCC to generate the subgoal sequences.
The agent completes these subgoals in the specified order to accomplish the goal task.
Formally, for each language encoding $\hat{x}_i$, the PCC predicts the subgoal action as:
\begin{equation}
s_i = \arg\max_k (FC_{1} (\hat{x}_i)), \;\;\; \text{where}\;k \in [1,N_{subgoals}],
\end{equation}
where $FC_{1}$ denotes a single layer perceptron, and $N_{subgoals}$ denotes the number of subgoals.
We train the PCC module using imitation learning, with the associated subgoal labels.
On the validation split used in \cite{shridhar2020alfred}, the controller achieves 98.5\% accuracy.
\rev{We provide further details on these subgoals in the supplementary for space's sake.}

\subsection{Master Policy}
\label{sec:master_policy}
As we discussed in Sec. `Model,' the reasoning required for navigation is significantly varied from the interaction. 
To this end, we propose to use a dedicated module for navigation, which we call  `master policy' (illustrated by the upper-right blue box in Fig. \ref{fig:overview}). 
It not only performs navigation but simultaneously also marks the locations for object interaction along the way. 
In other words, it generates the navigational action sequence based on the multi-modal inputs.

Specifically, let $\mathcal{A}_n$ denote the set of primitive navigation actions \{\textsc{MoveAhead}, \textsc{RotateRight}, \textsc{RotateLeft}, \textsc{LookUp}, \textsc{LookDown}\}.
The master policy learns to navigate in the environment by learning a distribution over $\mathcal{A}_n \cup$ \texttt{<MANIPULATE>}, where \texttt{<MANIPULATE>} is the abstract token we introduce for the agent to signify when to move control to the next level of the hierarchy, \ie, the interaction policies, for completing manipulation subgoals. 
It comprises of two modules: (1) an \textit{object encoding module} that provides information about the object the agent needs to locate for interaction, and (2) a navigation policy that outputs the navigation action sequence based on the multi-modal input for traversing the environment.
For the instruction to be used in the master policy, we additionally propose a new way for combining subtask language instructions.

\subsubsection{Subtask language encoding.}
\label{sec:lang_dist}

\rev{The language instructions for a given task can be divided into two types; (1) navigation and (2) interaction. We observed that for completing a given compositional task, the agent needs to navigate to the necessary locations and then interact with relevant objects.
An embodied task would consist of multiple combinations of such pairs with varying locations and interaction subgoals.} 

We further propose a method for encoding the combination of instructions for navigation.
In particular, we regard the subtask instruction as a combination of (1) navigation to discover the relevant object and (2) corresponding interactions. For instance, in the subtask, \textit{``Turn around and walk to the garbage bin by the TV. Pick up the blue credit card on the TV stand.''}, the agent needs to interact with the credit card, which is crucial information for the agent and also serves as a navigational criterion, \ie, the agent should stop if it encounters the credit card in close vicinity.
{We observe that this information is often missing in a navigation command but present in the next interaction instruction.
We encode these language instruction combinations in a similar manner as PCC.
Here, $\hat{l}_{T:T+1,n}$ refers to the encoded feature of the combined subtask instruction of the navigation subgoal $T$ and the corresponding interaction subgoal $T+1$.}

\subsubsection{Object encoding module (OEM) (box in orange).}
\label{sec:OEM}

Locating the required objects in an essential part of navigation.
Trying to interact with incorrect objects can lead to catastrophic failure.
To find the correct object, we propose an object encoding module that takes as input the subtask language instruction \rev{$l_{T:T+1,n}$} and gives the target object that the agent must locate for interaction.
This guides the agent's navigation by acting as a \textit{navigation subgoal monitor} that indicates the end of the navigation subgoal and shifts control to the next interaction policy.
The object encoder is composed of a Bi-LSTM with a two-layer perceptron which outputs the object class (Eq. \ref{eq:nav_act}).
During navigation, the subgoal monitor uses a pretrained object detector \cite{he2017mask} that validates if the relevant object is present in the current view or not.
If the agent spots the item, it switches to the appropriate interaction policy; otherwise, it continues to navigate.

\subsubsection{Navigation policy (box in yellow).}
\label{sec:Nav_Policy}
The second component of the master policy is the navigation policy that generates the sequence of navigable low-level actions using the processed multi-modal data as input. The architecture is based on the action prediction module of \cite{singh2021factorizing}.
It uses visual features, subtask instruction features, object encoding and the embedding of the preceding time step action as inputs.
The goal of the navigation policy is locating the correct object for interaction. Therefore, it utilises the subtask combination instruction $l_{T:T+1}$ as input which provides low-level information relevant for navigation as well as the information about the object that the agent needs to interact with.
This aids the agent in arriving at the correct location. 
To capture the relationship between the visual observation and language instructions, we dynamically generate filters based on the attended language features and convolve visual features with the filters, denoted by "VL-Ground" in Fig. \ref{fig:overview}.
To summarise, the LSTM hidden state $h_{t,n}$ of the master policy decoder, $\text{LSTM}_n$, is updated with four different features concatenated together as:
\begin{equation}
    \begin{split}
        o_t &= \operatorname*{argmax}_{k'}(\text{FC}_o(\hat{l}_{\rev{T:T+1},n})) \;\;\; \; k' \in [1,N_{objects}]\\
        h_{t,n} &= \text{LSTM}_n([\hat{v}^{pan}_{t}; \;\hat{l}_{\rev{T:T+1},n};\; a_{t-1,n};\; o_{t}])\\
        a_{t,n} &= \operatorname*{argmax}_k  (\text{FC}_{n}([\hat{v}^{pan}_{t}; \hat{l}_{\rev{T:T+1},n}; a_{t-1,n}; o_{t}; h_{t,n}]))\\
                & ~~~~~~~~~~~~~~~~~~~~~~~~~~~~~~~~~~~~~~~~~~~~ \text{where}~k \in [1, |\mathcal{A}_n|+1]
        \label{eq:nav_act}
    \end{split}
\end{equation}
where $\hat{v}^{pan}_{t}$ denotes the attended visual features for surrounding views (See supp.) at time step $t$; $\hat{l}_{\rev{T:T+1},n}$ the attended subtask language features for the navigation subgoal $T$ and the corresponding interaction subgoal $T+1$; $a_{t-1,n}$ the action given by master policy in the previous time step; and $o_t$ the object encoding given by the OEM.

\subsubsection{Loop escape (box in gray).}
\label{sec:lem}
In addition, we use subgoal progress monitor and overall progress monitor similar to \cite{shridhar2020alfred} to train the navigation policy \reva{and also utilize a heuristic loop escape module for escaping the deadlock conditions.}
We provide details in the supplementary.

\newcommand{\mcc}[2]{\multicolumn{#1}{c}{#2}}
\newcommand{\mcp}[2]{\multicolumn{#1}{c@{\hspace{30pt}}}{#2}}
\definecolor{Gray}{gray}{0.90}
\newcolumntype{a}{>{\columncolor{Gray}}r}
\newcolumntype{b}{>{\columncolor{Gray}}c}
\newcommand{\B}[1]{\textcolor{blue}{\textbf{#1}}}

\begin{table*}[t!]
    \centering
    \setlength{\tabcolsep}{2.2pt}
    \resizebox{\textwidth}{!}{
        \begin{tabular}{@{}lccccccccaarrcaarr@{}}
            \toprule
            
            \multirow{3}{*}{Model} & & \mcc{1}{\textbf{Language}} & & \mcc{4}{\textbf{Model}} & & \mcc{4}{\textbf{Validation}}  & & \mcc{4}{\textbf{Test}} \\
            
            \cmidrule{3-3} \cmidrule{5-8} \cmidrule{10-18}
            
            & & \multirow{2}{*}{Goal-Only} & & \mcc{1}{Rule-based} & & \mcc{1}{Semantic} & \mcc{1}{Subtask} 
            & & \mcc{2}{\textit{Seen}} & \mcc{2}{\textit{Unseen}} & & \mcc{2}{\textit{Seen}}   & \mcc{2}{\textit{Unseen}}  \\
            & &  & & \multicolumn{1}{c}{Planning} & & \multicolumn{1}{c}{Memory} & \multicolumn{1}{c}{Division} 
            & & \multicolumn{1}{b}{SR} & \multicolumn{1}{b}{PLWSR} & \multicolumn{1}{c}{SR} & \multicolumn{1}{c}{PLWSR}  
            & & \multicolumn{1}{b}{SR} & \multicolumn{1}{b}{PLWSR} & \multicolumn{1}{c}{SR} & \multicolumn{1}{c}{PLWSR} \\
                             
            \cmidrule{1-1} \cmidrule{3-3} \cmidrule{5-8} \cmidrule{10-18}

            Seq2Seq \cite{shridhar2020alfred}
            & & \xmark & & \xmark & & \xmark & -
            & & 3.70 & 2.10 & 0.00 & 0.00
            & & 3.98 & 2.02 & 0.39 & 0.80 \\
            MOCA \cite{singh2021factorizing}
            & & \xmark & & \xmark & & \xmark & -
            & & 25.85 & 18.95 & 5.36 & 3.19
            & & 26.81 &  19.52  & 7.65  & 4.21 \\
            EmBERT \cite{suglia2021embodied}
            & & \xmark & & \xmark & & \xmark & -
            & &{ 37.44} & \textbf{28.81} & 5.73 & 3.09
            & & 31.77  & 23.41 & 7.52  & 3.58  \\
            E.T. \cite{pashevich2021episodic}
            & & \xmark & & \xmark & & \xmark & -
            & & \textbf{46.59} & - & 7.32 & -
            & & \textbf{38.42} & \textbf{27.78} & 8.57  & 4.10  \\
            LWIT \cite{nguyen2021look}
            & & \xmark & & \xmark & & \xmark & -
            & & 33.70 & 28.40 & 9.70 & 7.30
            & & 30.92  & 25.90 & 9.42  & 5.60  \\
            HiTUT \cite{zhang2021hierarchical}
            & & \xmark & & \xmark & & \xmark & Subgoal
            & & 25.24 & 12.20 & 12.44 & 6.85
            & & 21.27  &  11.10 & 13.87  & 5.86 \\
            \reva{M-Track \cite{song2022one}}
            & & \xmark & & \xmark & & \xmark & Binary
            & &  26.70 & - & 17.29 & -
            & &  24.79 &  13.88  & 16.29 & 7.66 \\

            \midrule
            
            {\bf \method (Ours)}
            & & \xmark & & \xmark & & \xmark & Subgoal
            & & 34.39 & 23.04 & \textbf{20.08} & \textbf{10.84}
            & & 30.13 & 21.19 & \textbf{17.04} & \textbf{9.69} \\
            
            \midrule
            
            LAV \cite{nottingham2021modular}
            & & \cmark & & \cmark & & \xmark & Subgoal
            & & 12.70 & 5.9 & - & - 
            & & 13.35 & 6.31 & 6.38  & 3.12  \\

            HLSM \cite{blukis2021persistent}
            & & \cmark & & \xmark & & \cmark & Primitive
            & & 29.63 & - & {18.28} & -
            & & 29.94 & 8.74  & 20.27 & 5.55 \\

            \reva{MAT \cite{ishikawa2022moment}}
            & & \cmark & & \xmark & & \cmark & Primitive
            & & 30.98 & - & 17.66 & -
            & & 33.01 & - & 21.84 & - \\

            \reva{FILM \cite{min2021film}} 
            & & \xmark & & \cmark & & \cmark & Primitive 
            & & 38.51 & 15.06 & 27.67 & 11.23
            & & 27.67 & 11.23 & 26.49 & 10.55 \\
            
            \reva{EPA \cite{liu2022planning}}
            & & \cmark & & \cmark & & \cmark & -
            & & - & - & - & -
            & & 39.96 & 2.56  & 36.07 & 2.92 \\

            \bottomrule
        \end{tabular}
    }
    \vspace{-0.5em}
    \caption{Task and Goal-Condition Success Rates.
        {\cmark} in ``Goal-Only'' column under ``Language'' indicates that the corresponding approach uses only goal statements.
        ``Rule-based Planning'' indicates if a model exploits rule-based planning such as shortest path algorithms.
        ``Semantic Memory'' denotes if the approach requires external memory for storing semantic information (\eg, object positions, classes, \etc) using data structures (\eg, grid maps, graphs, \etc).
        ``Subtask Division'' represents if an agent breaks a task into subtasks (Primitive/Subgoal/Binary) or not (-).
        A subtask can be a ``Primitive'' interaction action, a set of ``Subgoal'' actions, or a ``Binary'' indicator for navigation/interaction.
        Our \method achieves the highest unseen SR and PLWSR in both validation and test folds compared to prior works without rule-based planning or semantic memories.
        We indicate the highest values in bold among them.
        \vspace{-1.5em}
    }
    \label{tab:sota_compare}
\end{table*}

\subsection{Interaction Policy}
\label{sec:IP}
To abstract a visual observation to a consequent action, the agent requires a global scene-level comprehension of the visual observation whereas, for the localisation task, the agent needs to focus on both global as well local object-specific information.
Following \cite{singh2021factorizing}, we exploit separate streams for action prediction and object localization due to the contrasting nature of the two tasks, illustrated as `Interaction Policy$_\text{k}$' in Fig. \ref{fig:overview}.
Each interaction policy consists of an action policy module which is responsible for predicting the sequence of actions corresponding to the interaction subgoal, and an interaction perception module which generates the pixel-level segmentation mask for objects that the agent needs to interact with at a particular time step.

The task requires the execution of varied subgoals with different levels of complexity.
For instance, a \textsc{Heat} subgoal might require interaction with either a stove or a microwave whereas for a \textsc{Pickup} subgoal, there is a variety of receptacles but the action sequence is simpler.
To focus on individual sub-objectives, we train an interaction policy, for each subgoal where $k \in [1, N_{subgoals}]$.
We observed that each interaction has its own properties and the navigation information history is irrelevant to the task, which allows us to keep an isolated hidden state for each interaction subgoal.
We provide further details about the architecture and the training process for interaction policies in the supplementary.

\section{Experiments}
\label{sec:results}

\noindent
\textbf{Dataset.}
To evaluate our approach in challenging scenarios, we focus on the problem of interactive instruction following in the ALFRED benchmark \cite{shridhar2020alfred}, which poses numerous challenges including long-term planning, partial observability, and irreversible state changes.
To complete a task successfully, an agent needs to navigate through very long horizons. Along the trajectory, the agent can interact with 118 objects in novel environments, which requires a thorough comprehension of both visual observations and their relation with the natural language directives.
It provides expert trajectories for the agents performing household tasks in simulated environments on AI2-THOR \cite{ai2thor}.
The dataset is divided into three splits; `train', `validation', and `test' set.
To evaluate the generalisation ability of an embodied agent to novel environments, the benchmark further divides 'validation' and 'test' trajectories into \textit{seen} and \textit{unseen} splits.
\textit{Unseen} comprises a set of rooms that are held out during training and scenes that are exposed to the agent during training are termed as \textit{seen}.
For each task, ALFRED provides a goal statement with multiple (4+) step-by-step instructions describing each subtask. (Supp. Sec. `Subgoal Evaluation').

\vspace{-0.2em}
\noindent\paragraph{Metrics.}
\reva{We use the widely used evaluation metrics in literature~\cite{shridhar2020alfred,padmakumar2022teach}, success rate (SR) is the ratio of the successfully completed episodes to the total episodes.}
The path length weighted success rate (PLWSR) penalizes the success rate by the length of the trajectory traversed by the agent, which indicates the efficiency of the embodied agent.
\reva{The goal-condition success rate (Goal-Cond.) is the ratio of the satisfied conditions among the total goal conditions for tasks, which takes into account the partial task completion ability of the agent.}

\subsection{Comparison with State of the Arts}
First, we conduct a quantitative analysis of task success rates (SR) and path length weighted success rates (PLWSR) \cite{anderson2018evaluation} by comparing our approach with prior arts on the interactive instruction following task \cite{shridhar2020alfred} and summarize the results in Table \ref{tab:sota_compare}.
\reva{We indicate the highest value for each metric in bold font among methods that are comparable to ours that do not use rule-based planning or semantic memories for a fair comparison.
We also present recent methods that use expensive external supervision or well-designed planners for reference.}

We observe that in unseen environments, \method outperforms \reva{most} prior-arts in terms of PLWSR for both test and validation folds. This demonstrates the ability of our agent to accomplish tasks in novel environments with higher efficiency. 
For seen environments in the test fold, \method shows comparable performance with LWIT and EmBERT in terms of SR and PLWSR but these works exhibit relatively stronger bias towards seen environments, which is evidenced by the significant drop in their unseen SR (\ie, 69.5\% and 76.3\% relative drop, respectively).
Similarly, E.T. decently performs in seen environments but shows a drastic drop (77.7\% relative) of SR in unseen environments. Note that E.T. utilises extra synthetic training data.

\begin{figure*}[t]
    \centering
    \begin{subfigure}{.42\linewidth}
        \centering
        \includegraphics[width=\linewidth]{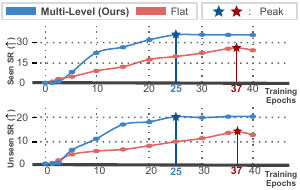}
        \vspace{-1.5em}
        \caption{Training efficiency}
        \label{fig:HvsF1}
    \end{subfigure}
    \hspace{4em}
    \begin{subfigure}{.42\linewidth}
        \centering
        \includegraphics[width=\linewidth]{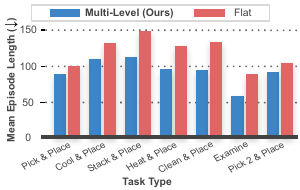}
        \vspace{-1.5em}
        \caption{Mean episode length}
        \label{fig:HvsF2}
    \end{subfigure}%
    \vspace{-0.5em}
    \caption{
        Multi-level policy learns faster and more effective action sequences.
        Plot (a) shows the learning curves (success rates vs. epochs) of the hierarchical and flat policy agents for unseen and seen environments.
        Plot (b) presents the average length of an episode traversed by a hierarchical or flat policy for the seven task types \cite{shridhar2020alfred}.
        The flat policy denotes the NIH ablated agent, \#(c) in Table \ref{tab:ablation}.
        \vspace{-0.5em}
    }
    \label{fig:HvsF}
    \vspace{-1em}
\end{figure*}

\subsection{Bias Towards Seen Environment}
\label{subsec:env_bias}

It is previously observed that embodied agents relying on low-level visual features for perception generally exhibit bias towards seen environments~\cite{zhang2020diagnosing}.
Unfortunately, \method also exhibits similar bias towards seen environments but the degree is significantly lower than other works (E.T., LAV, MOCA).

To mitigate this bias, recent works such as HLSM, MAT, FILM, and EPA utilize spatial semantic representations based on the depth followed by additional depth supervision and semantic segmentation data by assuming perfect egomotion that enables retrieval of accurate camera poses for estimation of the environment layout.
These assumptions limit the approaches' deployment capabilities since perfect egomotion may not be accessible in a real-world scenario and such spatial representations may lead to an exponential increase in memory requirements when deployed in larger environments.
Note that our approach outperforms all these works and shows comparable performance with FILM in terms of PLWSR score without requiring additional memory and perfect egomotion for generating spatial maps.

Furthermore, HLSM and MAT redefine the agent's action space to adopt a pretrained grid-based navigation system on 3D semantic maps for effective navigation.
Similarly, FILM and EPA are equipped with rule-based algorithms for obstacle-free path planning.
These agents incorporate heuristics for performance gains whereas \method utilises purely learning-based algorithms.
While the heuristics may help task completion (improved SR), they adversely affect the efficiency and generalisation of the agents as evidenced by the drop in unseen PLWSR for HLSM and EPA.

\subsection{Multi-Level Policy \vs Flat Policy}
\label{sec:hier_vs_flat}

We compare the learning efficiency for the hierarchical and flat policy agents for seen and unseen environments. The performance of our hierarchical agent and the flat agent are compared quantitatively in relation to the number of iterations (expressed in terms of epochs), and the results are presented in Fig. \ref{fig:HvsF}. As shown, the multi-level hierarchical policy gives a major improvement over the flat policy. Higher success rates in unseen scenarios evidence its ability to perform in novel environments. As depicted in Table \ref{tab:ablation} (\#(a) \textit{vs.} \#(c)), for seen and unseen task SR, the hierarchical agent outperforms the flat policy by 8.04\% and 7.65\%, respectively. In both seen and unseen ‘Goal-Cond.', the hierarchical approach outperforms, with improvements of 10.51\% and 9.38\%, respectively. The greater performance of the hierarchical approach on both overall task success rate and goal condition suggests its comprehension of both short-term subtasks and long-horizon whole tasks. 

The multi-level hierarchical agent converges significantly faster than the flat agent (25$^\text{th}$ epoch \textit{vs.} $37^\text{th}$ epoch), as shown in Fig. \ref{fig:HvsF1}, demonstrating the computational efficiency of our approach. 
Our policies are trained in two stages. 
We train interaction policies first, which collectively takes two epochs to converge.
We provide the details on the convergence of the interaction policies in the supplementary.
We include them in computation and begin the hierarchical agent's curve from the 3$^\text{rd}$ epoch, which is effectively the 1$^\text{st}$ epoch for the master policy.

Fig. \ref{fig:HvsF2} represents the average lengths of a successful trajectory traversed by the hierarchical and flat policy agents, for different task types, contrasting the efficiency of each agent.
The hierarchical agent consists of the master policy that is dedicated solely to navigation, giving it a significant advantage over the flat agent that learns everything using the same network parameters.
It was observed that due to the wide action space, the flat agent occasionally executes irrelevant interactions along the trajectory, which is not the case with \method.
The dedicated action sets for the master policy and interaction policies improve \method by allowing the agent to avoid any unnecessary interactions while traversing to discover the desired object.
The interaction policies also perform significantly better because they only master certain short-horizon tasks, which speeds up and simplifies the learning process.
We also provide the subgoal performance for each module in the supplementary.

\subsection{Interpretable Subgoals}
\label{subsec:interpretability}
\reva{The interpretability of embodied agents recently gain attention in the literature for the transparency of their reasoning process \cite{shivansh2021interpretation,kshitij2022what}.
Despite recent advances in the domain, many approaches still provide little to no transparency about the agent's actions due to their primitive action space that cannot fully represent the intention of the agents.
To demystify the agent's behavior, \method generates semantically meaningful subgoals that subject the agent's logic to observation (`What is the agent attempting to accomplish right now?').
This makes it easier for humans to monitor the progress of task.}

The generated subgoals are far more interpretable than low-level action sequences. 
For instance, a low-level \textsc{Put} action might be associated with any of the subgoals such as \textsc{Heat}, \textsc{Cool}, or \textsc{Put}.
The high-level semantics are reasoned about by the hierarchical agent, and the agent’s intent is considerably clearer. 
For instance, if the agent is performing a \textsc{Cool} subgoal, then it is more likely to interact with the refrigerator.
If it is a \textsc{Heat} subgoal, then it is more likely to interact with a microwave or stove. The subgoal information provided by the PCC provides extra useful information to the multi-level agent as well as the observer.
In contrast, the flat policy agent considers it as a single atomic action regardless of the object or receptacle involved.

\subsection{Ablation Study}
\label{sec:ablation}

We conduct a series of ablation analyses on the \reva{proposed} components of \method and report the results in Table \ref{tab:ablation} to evaluate the significance of each module.
\reva{In the supplementary, we further provide ablation studies for model input, design components, task types, and subgoal types}.

\begin{table}[t!]
    \centering
    \setlength{\tabcolsep}{1.4pt}
    \resizebox{\columnwidth}{!}{
        \begin{tabular}{@{}cccccccccccc@{}}
            \toprule
            & \multicolumn{3}{c}{\bf Components} && \multicolumn{2}{c}{\bf Validation-Seen} && \multicolumn{2}{c}{\bf Validation-Unseen} \\
            \cmidrule(r){1-1} \cmidrule(lr){2-4} \cmidrule{6-7} \cmidrule{9-10}
            
            \# & \begin{tabular}{c}MIP\end{tabular}  & \begin{tabular}{c}NIH\end{tabular} & \begin{tabular}{c}OEM\end{tabular}  & & \multirow{1}{*}{Task}   & \multirow{1}{*}{Goal-Cond.} & & \multirow{1}{*}{Task} &  \multirow{1}{*}{Goal-Cond.} \\
            
            \cmidrule(r){1-1} \cmidrule(lr){2-4} \cmidrule{6-7} \cmidrule{9-10}
            a) & \cmark & \cmark  & \cmark &&$34.39_{(0.2)}$	&$41.96_{(0.5)}$	&&$20.08_{(0.3)}$	&$38.42_{(0.2)}$\\
            \cmidrule(r){1-1} \cmidrule(lr){2-4} \cmidrule{6-7} \cmidrule{9-10}
            
            b) & \cmark & \cmark & \xmark &&$28.61_{(0.4)}$	&$32.96_{(0.3)}$	&&$13.31_{(0.9)}$	&$29.13_{(0.3)}$ \\
            c) & \cmark & \xmark & \cmark & &$26.35_{(0.9)}$	&$31.45_{(0.9)}$	&&$12.43_{(0.7)}$	&$29.04_{(0.9)}$\\
            
            d) & \xmark & \cmark & \cmark & &$31.08_{(0.9)}$	&$39.26_{(0.8)}$	&&$15.82_{(0.5)}$	&$30.81_{(0.3)}$\\
            
            e) & \xmark & \xmark & \cmark & &$20.59_{(1.4)}$	&$25.13_{(2.1)}$	&& \hspace{5pt}$7.45_{(1.1)}$	&$14.05_{(1.0)}$\\
            
            f) & \xmark & \cmark & \xmark & &$23.54_{(1.8)}$	&$31.61_{(1.5)}$	&&$10.30_{(0.7)}$	&$25.42_{(1.1)}$\\
            \bottomrule
        \end{tabular}
    }
    \vspace{-0.5em}
    \caption{
        Ablation study for components of \method.
        \reva{
        We report the task success rate for each ablation.
        {\cmark} and {\xmark} denote that the corresponding component is {\color{black}{present}}/{\color{black}{absent}} in \method.
        MIP (Modular Interaction Policy) denotes the subgoal modules for interaction policies.
        NIH (Navigation Interaction Hierarchy) denotes the third level of hierarchy between navigation and interaction policies.
        OEM (Object Encoding Module) denotes the object encoding module.
        We report averages of 5 runs with random seeds with standard deviations depicted in sub-script parentheses (\eg, (0.2)).
        }
        \vspace{-0.5em}
    }
    \vspace{-1em}
    \label{tab:ablation}
\end{table}

\subsubsection{Without object encoding module (OEM).}
We ablate the navigation subgoal monitor and train the navigation policy without object information. The agent can complete some objectives, but it lacks object information, which functions as a stopping criterion, preventing proper navigation. Hence, it is unable to completely comprehend the relationship between the step-by-step instruction and the visual trajectory. This limits the agent's capacity to explore and connect various interaction policies required for task completion, leading to a significant performance drop (Table \ref{tab:ablation} \#{(a)} \textit{vs.} \#{(b)}).

\subsubsection{Without navigation interaction hierarchy (NIH).}
Next, we demonstrate the importance of hierarchy between navigation and interaction policies \ie the second level of hierarchy in our framework. For this, we utilize the same network for learning navigation and interaction action prediction.
For interaction mask generation, we preserve the interaction perception module. To ablate the benefits of the \reva{subtask language encoding}, we use the concatenation of all step-by-step instructions as language input and conduct action and mask prediction while leaving the other modules unaltered. The ablated model's task success rates drop significantly (Table \ref{tab:ablation} \#(a) \textit{vs.} \#(c)), showing that it is unable to effectively utilise the available inputs.

\subsubsection{Without modular interaction policy (MIP).}
\label{sec:mod_vs_nonmod}
In modular networks, the decision-making process is separated into numerous modules. Each module is designed to perform a certain function and is put together in a structure that is unique to each trajectory instance. Because of their compositional nature, such networks with the help of specialised modules often perform better in new environments than their flat counterparts \cite{hu2019you,blukis2019learning}.
We present a quantitative comparison of interaction policies' modular structure \rev{(Table \ref{tab:ablation} \#(a) \textit{vs.} \#(d))}. For this experiment, we train a single policy module to learn all  interaction tasks. The decoupled pipeline for action and mask prediction, as well as the rest of the settings, are preserved. The modular agent outperforms the non-modular agent by \rev{3.31\%} and \rev{4.26\%} in seen and unseen task SR, respectively. It also performs significantly well in both seen and unseen 'Goal-Cond.' criteria, with gains of \rev{2.70\%} and \rev{7.61\%}, respectively. The greater performance of the modular policy in both task and goal-condition metrics highlights the \rev{benefits of modular structure in long-horizon planning tasks}.
Next, we provide the individual performance of the two major components of our framework which brings the most empirical gain, OEM and NIH in the absence of other components.

\subsubsection{Object encoding module only.} 
In this ablation, we evaluate the effect of the object encoding module (OEM) in the absence of the hierarchical and modular structure.
This makes the agent flat and thus analogous to \cite{singh2021factorizing} except for OEM. The agent (Table \ref{tab:ablation} \#(e)) demonstrates significantly higher performance than \cite{singh2021factorizing}, highlighting the relevance of target object information for navigation and the effectiveness of the proposed OEM.

\subsubsection{Navigation interaction hierarchy only.}
While ablating the modular structure and object encoding module, we observe  degradation in performance (Table \ref{tab:ablation} \#(f)) which implies that the multi-level hierarchical architecture needs to include other proposed components for optimal performance. The overall performance improves when these components are combined (\#(a) \textit{vs.} \#(f)) indicating that the proposed components are complementary to each other.

\section{Conclusion}
We address the problem of interactive instruction following.
To effectively tackle the long horizon task, we propose a multi-level compositional approach to learn agents that navigate and manipulate objects in a divide-and-conquer manner for the diverse nature of the entailing task.
To improve navigation performance, we propose an object encoding module to explicitly encode target object information during internal state updates.
Our approach yields competitive performance with higher efficiency than prior arts in novel environments without extra supervision and well-designed planners.

\section*{Acknowledgments}
This work is partly supported by the NRF grant (No.2022R1A2C4002300), IITP grants (No.2020-0-01361-003, AI Graduate School Program (Yonsei University) 5\%, No.2021-0-02068, AI Innovation Hub 5\%, 2022-0-00077, 15\%, 2022-0-00113, 15\%, 2022-0-00959, 15\%, 2022-0-00871, 20\%, 2022-0-00951, 20\%) funded by the Korea government (MSIT).

\bibliography{aaai23}

\clearpage
\appendix

\section{Supplemenatry Materials for \\ Multi-Level Compositional Reasoning for Interactive Instruction Following}
~
~
~

\definecolor{Gray}{gray}{0.90}
\newcolumntype{a}{>{\columncolor{Gray}}r}
\newcolumntype{b}{>{\columncolor{Gray}}c}
\section{Approach Details}

\subsection{Surrounding Perception}
\label{sec:SP}
\method analyzes the surrounding views to improve the agent's perception and gain a better comprehension of the environment.
Let $v^d_t$ be a visual feature from each navigable direction, $d \in [0,D]$, at the time step, $t$, where $D$ represents the number of navigable directions and $d = 0$ the agent's egocentric direction.
After gathering additional observations, $v^{0:D}_t$, we acquire attended visual features, $\hat{v}^{pan}_t$, using the dynamic filters \cite{shridhar2020alfred} and concatenate all attended visual features as: 
\begin{equation}
    \begin{split}
        \hat{v}^{pan}_{t} &= [\hat{v}^0_{t}; \hat{v}^1_{t}; \cdots; \hat{v}^D_{t}].
        \label{eq:df_surrounding}
    \end{split}
\end{equation}
Using the attended visual features, we update the master policy decoder and predict actions as described in Eq. \ref{eq:nav_act}.

\begin{figure}[t!]
    \centering
    \begin{subfigure}{.9\columnwidth}
        \centering
        \includegraphics[width=\linewidth]{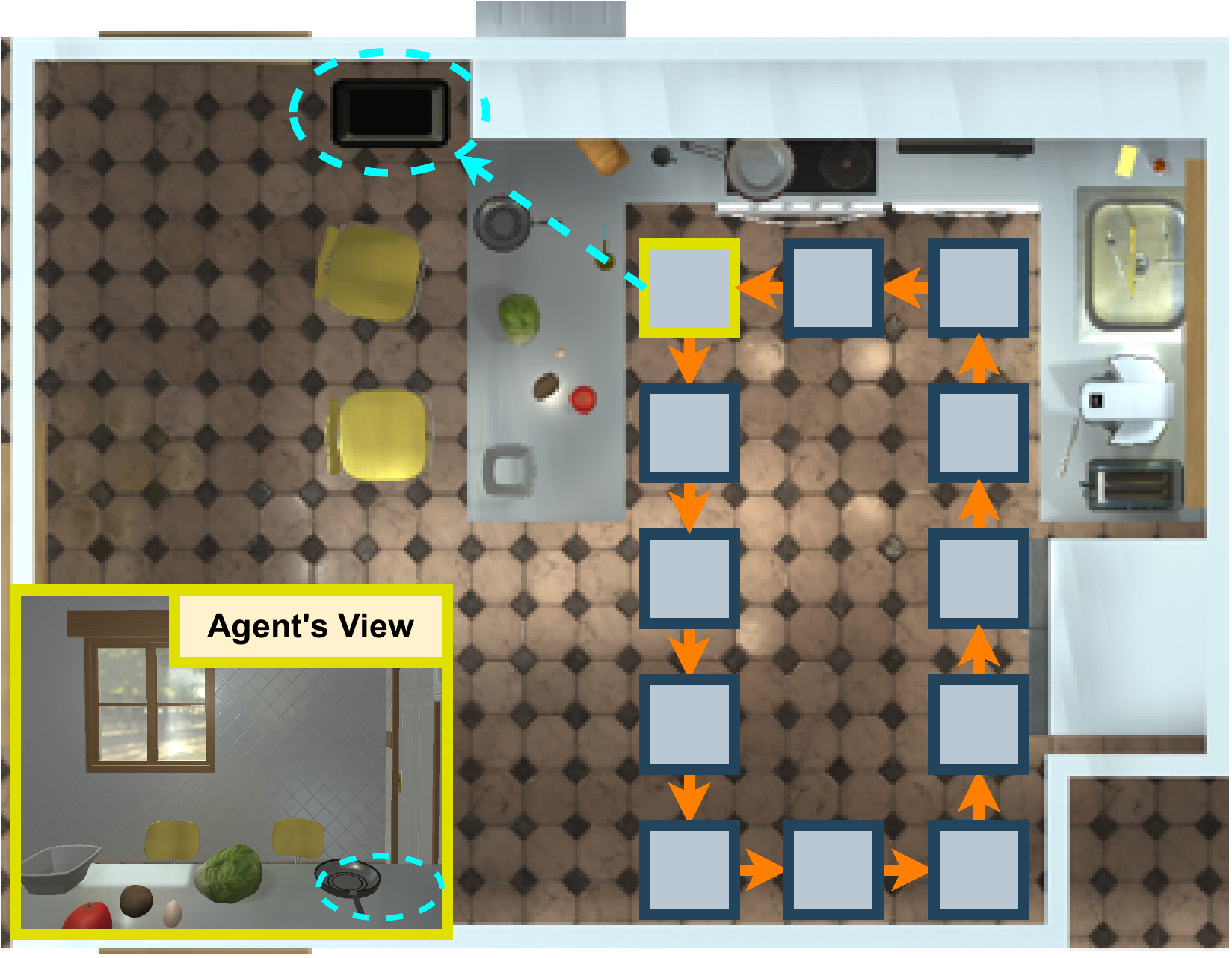}
        \caption{Deadlock State}
        \label{fig:stochastic_exploration_1}
    \end{subfigure}
    \begin{subfigure}{.9\columnwidth}
        \centering
        \includegraphics[width=\linewidth]{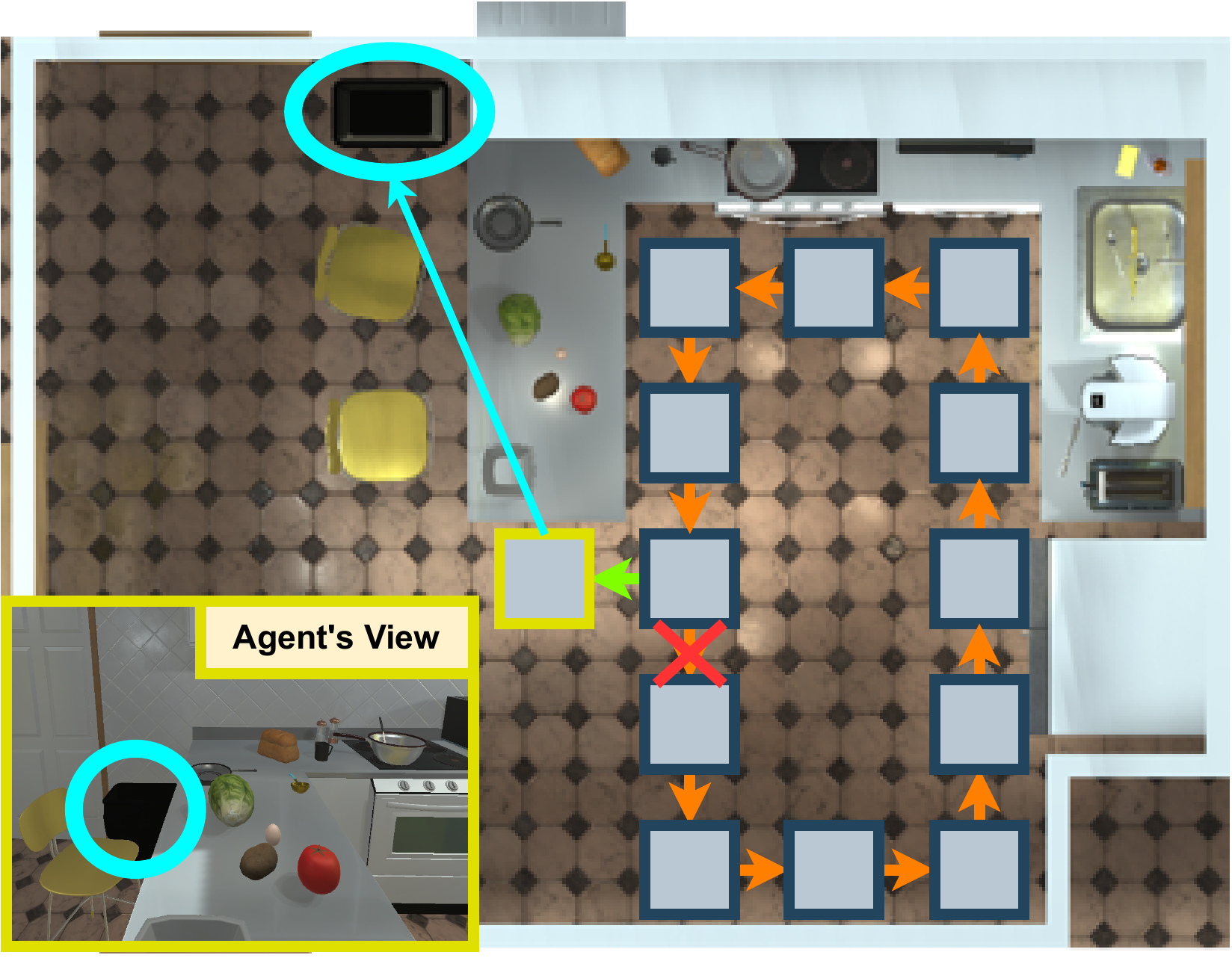}
        \caption{Loop Escape Module}
        \label{fig:stochastic_exploration_2}
    \end{subfigure}%
    \caption{
        \textbf{Loop escape module (LEM) for escaping deadlock states.}
        The objective of the agent at the current time step is to move to a target object (a garbage can).
        Figure (a) and (b) show an example of a deadlock state and the behavior of the loop escape module when finding the target object.
        Each dark-blue square denotes the position of the agent.
        {\color{cyan} $\circ$} denotes the target object that the agent should navigate to.
        {\color{cyan} $\xrightarrow{}$} denotes the view direction of the agent.
        The dashed {\color{cyan} $\circ$} and {\color{cyan} $\xrightarrow{}$} indicate that the target object is invisible to the agent due to occlusion.
        {\color{orange} $\xrightarrow{}$} denotes actions taken by the agent in a deadlock state.
        The loop escape module cancels the current action that causes the deadlock state, denoted by {\color{red} \bf $\times$}, and takes a stochastic action, denoted by {\color{green} $\xrightarrow{}$}.
    }
    \vspace{-1.5em}
    \label{fig:stochastic_exploration}
\end{figure}

\subsection{Loop Escape Module}
\label{sec:SE}
While navigating through an environment, the agent is prone to be caught around unanticipated obstructions such as tables or chairs, which might cause navigation failure. To circumvent such obstructions during navigation, \cite{shridhar2020alfred} introduced obstruction evasion (OE).
However, in the scenario where the agent predicts a repeated series of activities (i.e., a deadlock state), the agent may fail to navigate even without such barriers, resulting in limited exploration.
The OE mechanism is unable to handle such challenges since repeated sequences of visual observations might not be recognised as obstacles based on visual similarity in consecutive frames.
To address the problem, we propose \reva{a heuristic approach}, a loop escape module, which allows for improved exploration at inference time by avoiding deadlock conditions. Inspired by \cite{du2020learning}, we use external memory to memorise the history of visual observations.
The agent identifies a deadlock state if any $t'$ exists that satisfies Eq. \ref{eq:de}, given a history of visual observations, $v_{0:t}$, up to the present time step, $t$.
\begin{equation}
    \begin{split}
        v_{t' + w} = v_{t + w - W}, ~~ \forall w \in \; [1, W],
        \label{eq:de}
    \end{split}
\end{equation}
where $W$ indicates the length of the repeated actions.
In the event of a deadlock, the stochastic policy executes a random navigation action to assist the agent in escaping.
We observe our LEM is not sensitive to the hyperparameter $W$ by measuring the agent's performances across different $W$. We provide the sensitivity analysis of $W$ in the experiments. 

Fig. \ref{fig:stochastic_exploration} depicts an example of a deadlock state and the behaviour of the loop escape module in the case when the agent is unable to find a target object (a garbage can) due to occlusion (by the counter-top).
The agent in a deadlock state tries to revisit previously explored areas, which may limit exploration.
The loop escape module assists the agent in escaping the impasse condition, thus finding the garbage can.

\subsection{Interaction Policy}
\label{append:IP}
Let $\mathcal{A}_m$ denote the set of primitive interaction actions \{\textsc{Pickup}, \textsc{Put}, \textsc{Open}, \textsc{Close}, \textsc{ToggleOn}, \textsc{ToggleOff}\}.
The APM architecture consists of an LSTM decoder that takes as input the attended language encoding $\hat{l}_{\rev{T},m}$ corresponding to the current manipulation subgoal, the egocentric visual features $\hat{v}^{ego}_{t,m}$ and the previous time step action $a_{t,m}$ and outputs a distribution over the manipulation action space $\mathcal{A}_m \cup \texttt{<STOP>}$.
The manipulation policy uses $\texttt{<STOP>}$ token as an indicator for completion of the interaction task and shifts the control to the master policy for further task completion. 
Formally:
\begin{equation}
    \begin{split}
        h^{apm}_{t,m} &= \text{LSTM}_m ([\hat{v}^{ego}_{t,m}; \hat{l}_{\rev{T},m}; a_{t-1,m}]),\\
        a_{t,m} &= \operatorname*{argmax}_k  (\text{FC}_{m}([\hat{v}^{ego}_{t}; \hat{l}_{\rev{T},m}; a_{t-1,m}; h^{apm}_{t,m}]))\\
                & ~~~~~~~~~~~~~~~~~~~~~~~~~~~~~~~~~ \text{where} \;  k \in [1, |\mathcal{A}_m|+1].
    \end{split}
\end{equation}

The $\text{IPM}$ uses the same inputs as $\text{APM}$ and outputs the object class that the agent needs to interact with at the current time step.
Similar to \cite{shridhar2020alfred}, we use a pretrained Mask-RCNN to obtain the interaction mask corresponding to the object class.
For more details about the architecture, kindly refer to \cite{singh2021factorizing}.

\begin{equation}
    \begin{split}
        h^{vpm}_{t,v} &= \text{LSTM}_v ([\hat{v}^{ego}_{t,v}; \hat{l}_{\rev{T},v}; a_{t-1,v}]),\\
        c_{t,v} &= \operatorname*{argmax}_k  (\text{FC}_{v}([\hat{v}^{ego}_{t}; \hat{l}_{\rev{T},v}; a_{t-1,v}; h^{vpm}_{t,v}]))\\
                & ~~~~~~~~~~~~~~~~~~~~~~~~~~~~~~~~~ \text{where} \;  k \in [1, N_{objects}].
    \end{split}
\end{equation}

\rev{
We train our interaction policies in a supervised manner using imitation learning from expert ground truth trajectories.
Specifically, each interaction policy is trained to minimize the cross entropy loss as:
\begin{equation}
    \begin{split}
        \mathcal{L}_{int} = \sum_t a^*_{t,m} \log p_m + \sum_t c^*_{t,v} \log p_v,
    \end{split}
    \label{eq:int_loss}
\end{equation}
where $p_m = \text{FC}_{m}([\hat{v}^{ego}_{t}; \hat{l}_{\rev{T},m}; a_{t-1,m}; h^{apm}_{t,m}])$ and $p_v = \text{FC}_{v}([\hat{v}^{ego}_{t}; \hat{l}_{\rev{T},v}; a_{t-1,v}; h^{vpm}_{t,v}])$ denotes the probability distribution of action and class prediction, respectively, and $a^*_{t,m}$ and $c^*_{t,v}$ the corresponding ground truth actions and classes obtained from given expert trajectories, respectively.}

\section{Extended Experiments} 
\label{sec:experiments}

\subsection{Dataset and Metrics}
For training and evaluation, we used the ALFRED benchmark \cite{shridhar2020alfred}.
This benchmark provides expert trajectories for the agents performing household tasks (\eg, ``Put a knife in the table'') in a simulated environment on AI2-THOR \cite{ai2thor}.
The dataset is divided into three parts; `train', `validation', and `test' set. To evaluate the generalisation ability of an embodied agent to novel environments, the benchmark further divides 'validation' and 'test' trajectories into \textit{seen} and \textit{unseen} splits.
\textit{Unseen} comprises a set of rooms that are held out during training and scenes that are exposed to the agent during training are termed as \textit{seen}. 

The benchmark provides a high-level goal statement describing the final task and several low-level step-by-step instructions describing each subgoal that needs to be accomplished in the given order for successful completion of the goal task.
The benchmark has 7 types of subgoals; \textsc{GotoLocation}, \textsc{PickupObject}, \textsc{PutObject}, \textsc{CoolObject}, \textsc{HeatObject}, \textsc{CleanObject}, \textsc{SliceObject}, and \textsc{ToggleObject}.
\textsc{GotoLocation} denotes navigation subgoal where the agent requires to locate a target object for interaction.
\textsc{PickupObject} requires the agent to pick up an object.
\textsc{PutObject} requires the agent to put an object onto a receptacle.
\textsc{CoolObject}, \textsc{HeatObject}, and \textsc{CleanObject} require the agent to cool, heat, and clean an object while interacting with the required receptacle (\eg, fridge, microwave, sink basin, etc.).
\textsc{SliceObject} requires the agent to slice an object using a knife, resulting in sliced objects.
\textsc{ToggleObject} requires the agent to turn on or off an object such as a lamp.

\reva{For evaluation metrics, the success rate (SR) represents the ratio of the episodes that the agent successfully completes among the total episodes.}
The success rate penalized by the episode length of the agent (PLWSR) penalizes the success rate by the length of the trajectory traversed by the agent, which indicates the efficiency of an embodied agent and becomes one of the most important metrics in the benchmark.
\reva{Finally, the goal-condition success rate (Goal-Cond.) denotes the ratio of the satisfied conditions among the total goal conditions for tasks, which takes into account the partial task completion ability of the agent.}

\begin{table*}[t]
    \centering
    \setlength{\tabcolsep}{12pt}
    \resizebox{\linewidth}{!}{
        \begin{tabular}{@{}lacacacacac@{}}
            \toprule
            \multirow{2}{*}{Task-type}
                             & \multicolumn{2}{c}{Seq2Seq} & \multicolumn{2}{c}{Flat policy} & \multicolumn{2}{c}{\rev{HiTUT}} & \multicolumn{2}{c}{\rev{HLSM}} & \multicolumn{2}{c}{\method (Ours)} 
                             \\
                             \cmidrule(lr){2-3} \cmidrule(lr){4-5} \cmidrule(lr){6-7} \cmidrule(lr){8-9} \cmidrule(lr){10-11}
                             & \multicolumn{1}{a}{Seen} & \multicolumn{1}{c}{Unseen} 
                             & \multicolumn{1}{a}{Seen} & \multicolumn{1}{c}{Unseen} 
                             & \multicolumn{1}{a}{\rev{Seen}} & \multicolumn{1}{c}{\rev{Unseen}}
                             & \multicolumn{1}{a}{\rev{Seen}} & \multicolumn{1}{c}{\rev{Unseen}}
                             & \multicolumn{1}{a}{Seen} & \multicolumn{1}{c}{Unseen}
                             \\
                             
            \midrule

            {Pick     \& Place}  & $7.0$ & $0.0$ & $36.1$ & $5.0$  & \rev{$35.9$} & \rev{$26.0$} & \rev{$\B{57.0}$} & \rev{$\B{34.8}$}  & ${47.2}$ & ${9.0}$ \\
            {Cool     \& Place}  & $4.0$ & $0.0$ & $26.1$ & $19.1$ & \rev{$19.0$} & \rev{$4.6$}  & \rev{$17.5$} & \rev{$14.8$}            & $\B{34.1}$ & $\B{31.2}$ \\
            {Stack    \& Place}  & $0.9$ & $0.0$ & $13.9$ & $7.9$ & \rev{$12.2$} & \rev{$7.3$}  & \rev{$13.0$}  & \rev{${4.4}$}          & $\B{18.2}$ & $\B{12.8}$ \\
            {Heat     \& Place}  & $1.9$ & $0.0$ & $30.1$ & $10.5$ & \rev{$14.0$} & \rev{$11.9$} & \rev{$9.3$} & \rev{$0.0$}            & $\B{39.2}$ & $\B{17.0}$ \\
            {Clean    \& Place}  & $1.8$ & $0.0$ & $25.2$ & $22.1$ & \rev{$\B{50.0}$} & \rev{$21.2$} & \rev{$25.0$} & \rev{$11.3$} & ${33.0}$ & $\B{37.2}$ \\
            {Examine}            & $9.6$ & $0.0$ & $30.2$ & $8.4$ & \rev{$26.6$} & \rev{$8.1$} & \rev{$\B{46.8}$} & \rev{$\B{36.6}$} & ${39.4}$ & ${13.9}$ \\
            {Pick Two \& Place}  & $0.8$ & $0.0$ & $22.8$ & $14.4$ & \rev{$17.7$} & \rev{$12.4$} & \rev{$\B{34.7}$} & \rev{$18.0$} & ${29.8}$ & $\B{23.4}$ \\

            \midrule

            {Average}            & $3.7$ & $0.0$ & $26.3$ & $12.5$ & \rev{$25.1$} & \rev{$13.1$} & \rev{$29.0$} & \rev{$17.1$} & $\B{34.4}$ & $\B{20.6}$ \\

            \bottomrule
        \end{tabular}
    }
    \vspace{-0.5em}
    \caption{
        \textbf{Success rates across seven task types in ALFRED.}
        `Seq2Seq' refers to the method by \cite{shridhar2020alfred}.
        `Flat policy' refers to \textit{Without navigation interaction hierarchy (NIH)} in the `Ablation Study' section in the main paper.
        \rev{HiTUT \cite{zhang2021hierarchical} and HLSM \cite{blukis2021persistent} denote the prior arts involving hierarchical approach.}
        The highest values corresponding to each task type are in \B{blue} and all numbers are in percentage.
        We evaluate the agents on the validation set.
    }
    \vspace{-1.5em}
    \label{tab:task_type}
\end{table*}

\subsection{Implementation Details}
\label{supp:subsec:implementation}
Visual features are encoded with a pretrained ResNet-18 \cite{he2016deep} after resizing input images to $224\times224$.
To perceive surrounding views, we additionally gather visual observations from four navigable directions (i.e., \textsc{RotateLeft}, \textsc{RotateRight}, \textsc{LookUp}, and \textsc{LookDown}).
We train each module of our approach independently using the adam optimizer with an initial learning rate of $10^{-3}$.
We augment visual features by shuffling the channel order of each image \cite{shridhar2020alfred} and applying predefined image operations \cite{cubuk2018autoaugment}.
We set the coefficients for the subgoal and overall progress monitor objectives as 0.2 and 0.3.
\rev{The value of $W$ for the loop escape module is set as 10 using a grid search on the validation set.}

\begin{figure}[t!]
    \centering
    \includegraphics[width=\linewidth]{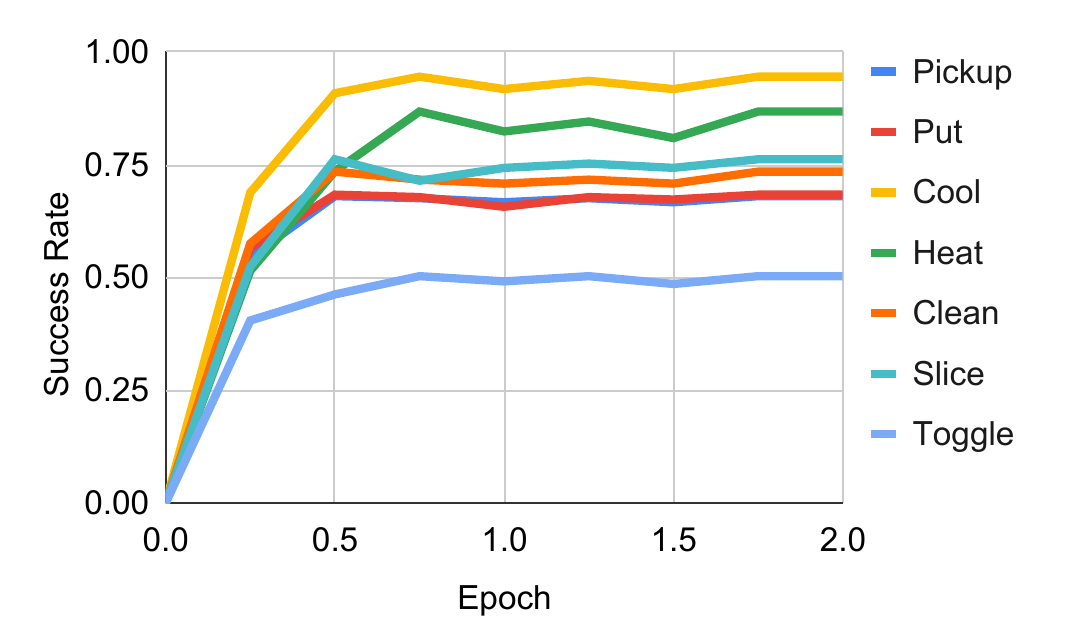}
    \caption{
        \textbf{Learning curves of subgoal policies.}
        The figure provides the learning curves for the subgoal policy training as discussed in Sec. `Training and Evaluation.'
    }
    \label{fig:subgaol_training}
\end{figure}

\subsection{Training and Evaluation}
\label{supp:sebsec:training}
We assess the performance of each subgoal policy in relation to its specific task.
On the ALFRED validation set, we present the success rate for each subgoal in Table \ref{tab:subgoal}.
To accomplish this, we employ the expert trajectory to guide the agent through the episode until it reaches the subgoal task.
The agent then uses the task-specific policy to infer the action sequence based on the language directive and visual observations.
The subgoal policies are trained using imitation learning, which collectively takes two epochs for convergence (Fig. \ref{fig:subgaol_training}).

Following that, we train and evaluate the master policy's performance using independent behaviour cloning on expert trajectories (\ie, assuming oracle subgoal-policies).
With ground truth subgoal sequences, the master policy is able to generalise effectively and achieve a 30.24\% success rate (SR) on unseen validation data. 
Then we train the object encoding module and the policy composition controller in isolation and combine all modules as described in Sec. `Model.'

\begin{table}[t!]
    \centering
    \setlength{\tabcolsep}{5.5pt}
    \resizebox{\columnwidth}{!}{
        \begin{tabular}{@{}lbcbcbcbc@{}}
            \toprule
            \multirow{2}{*}{Task-type}
                             & \multicolumn{2}{c}{Seq2Seq} & \multicolumn{2}{c}{Flat policy}& \multicolumn{2}{c}{\method(Ours)}
                             \\
                            \cmidrule(lr){2-3} \cmidrule(lr){4-5}
                            \cmidrule(lr){6-7}
                             & \multicolumn{1}{a}{Seen} & \multicolumn{1}{c}{Unseen} 
                             & \multicolumn{1}{a}{Seen} & \multicolumn{1}{c}{Unseen}
                             & \multicolumn{1}{a}{Seen} & \multicolumn{1}{c}{Unseen}
                             \\
                             
            \midrule

            {Pickup} & $32$ & $21$ & 66 & 39 & $\B{82}$ & $\B{68}$ \\
            {Put}    & ${81}$ & ${46}$ & 73 & 45 & $\B{90}$ & $\B{68}$ \\
            {Cool}   & ${88}$ & ${92}$ & 77 & 90 &  $\B{95}$ & $\B{94}$ \\
            {Heat}   & ${85}$ & ${89}$ & 69 & 49 &  $\B{85}$ & $\B{89}$ \\
            {Clean}  & ${81}$ & $57$ & 65 & 51 &  $\B{81}$ & $\B{73}$ \\
            {Slice}  & $25$ & $12$ & 65 & 49 &  $\B{73}$ & $\B{79}$ \\
            {Toggle} & ${100}$ & ${50}$ & 79 & 30 &  $\B{97}$ & $\B{53}$ \\

            \midrule
            {Average}    & $70$ & $46$ & 71 & 50 & $\B{86}$ & $\B{75}$ \\
            \bottomrule
        \end{tabular}
    }
    \caption{
        \textbf{Subgoal success rates.}
        The highest values per fold and task are in \B{blue}.
    }
    \label{tab:subgoal}
\end{table}

\begin{table*}[t!]
    \centering
    \setlength{\tabcolsep}{5pt}
    \resizebox{.9\textwidth}{!}{
        \begin{tabular}{@{}cccccccccc@{}}
            \toprule
            & \multicolumn{3}{c}{\bf Components} && \multicolumn{2}{c}{\bf Validation-Seen} && \multicolumn{2}{c}{\bf Validation-Unseen} \\
            \cmidrule(r){1-1} \cmidrule(lr){2-4} \cmidrule{6-7} \cmidrule{9-10}
           
            \# & \begin{tabular}{c} \rev{Object-centric} \\ \rev{mask prediction} \end{tabular} & \begin{tabular}{c} \rev{Factorised} \\ \rev{interaction policy} \end{tabular} & \begin{tabular}{c} \rev{Data} \\ \rev{Augmentation} \end{tabular} && \multirow{1}{*}{Task}   & \multirow{1}{*}{Goal-Cond.} &&  \multirow{1}{*}{Task} &  \multirow{1}{*}{Goal-Cond.} \\
            
            \cmidrule(r){1-1} \cmidrule(lr){2-4} \cmidrule{6-7} \cmidrule{9-10}
            - & \cmark & \cmark  & \cmark &&$34.39_{(0.25)}$	&$41.96_{(0.1)}$	&&$20.08_{(0.29)}$	&$38.42_{(0.15)}$\\
            \cmidrule(r){1-1} \cmidrule(lr){2-4} \cmidrule{6-7} \cmidrule{9-10}
            (a) & \cmark & \cmark & \xmark &&$32.00_{(0.96)}$	&$38.16_{(0.55)}$	&&$17.09_{(0.4)}$	&$34.62_{(0.13)}$\\
            (b) & \cmark & \xmark & \cmark &&$20.35_{(0.43)}$	&$29.61_{(0.33)}$	&&$11.02_{(0.28)}$	&$33.26_{(0.18)}$\\
            (c) & \xmark & \cmark & \cmark && \hspace{4pt} $6.90_{(0.19)}$	&$16.12_{(0.14)}$	&& \hspace{4pt} $2.51_{(0.21)}$	&$22.63_{(2.40)}$ \\
            
            \bottomrule
        \end{tabular}
    }
    \caption{
        \rev{
        \textbf{Ablation study for design components of \method.}
        We report the success rate for each metric.
        {\cmark}/{\xmark} denotes that a corresponding component is present/absent.
        Please refer to Sec. `Design Component Ablation' for details about each component.
        We report averages over 5 random runs with standard deviations in sub-script parentheses.
    }}
    \label{tab:supp_ablation}
\end{table*}

\subsection{Subgoal Evaluation}
\begin{figure*}[t!]
    \centering
    \begin{subfigure}{.8\linewidth}
        \centering
        \includegraphics[width=\linewidth]{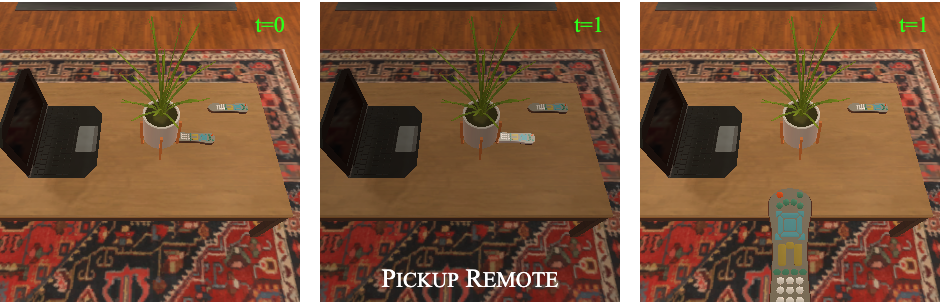}
    \end{subfigure}
    \caption{\textbf{Pickup Object} interaction policy agent completes a subgoal task `Take the remote.' in an unseen environment.}
    \label{fig:pickup}
    \vspace{-1.5em}
\end{figure*}

\begin{figure*}[t!]
    \centering
    \begin{subfigure}{.8\linewidth}
        \centering
        \includegraphics[width=\linewidth]{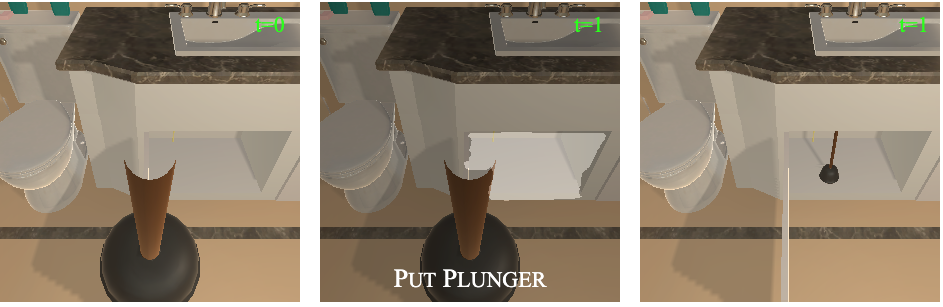}
    \end{subfigure}
    \caption{\textbf{Put Object} interaction policy agent completes a subgoal task “Put the plunger below the sink.” in an unseen environment.}
    \label{fig:put}
\end{figure*}

\begin{figure*}[t!]
    \centering
    \begin{subfigure}{0.8\linewidth}
        \centering
        \includegraphics[width=\linewidth]{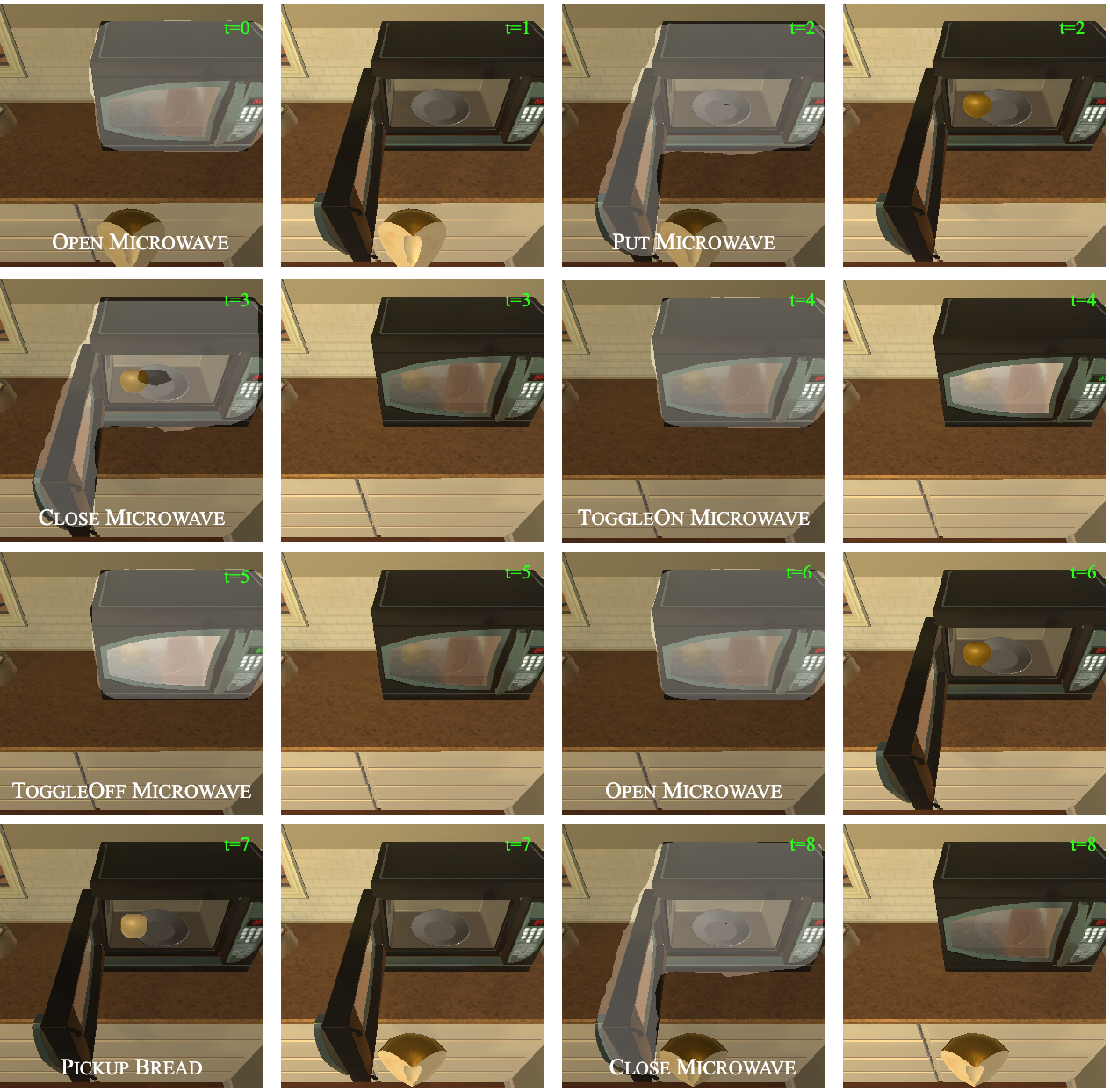}
    \end{subfigure}
    \caption{\textbf{Heat Object} interaction policy agent completes a subgoal task “Heat the apple slice in the microwave.” in an unseen environment.}
    \label{fig:heat}
\end{figure*}

\begin{figure*}[t!]
    \centering
    \begin{subfigure}{0.8\linewidth}
        \centering
        \includegraphics[width=\linewidth]{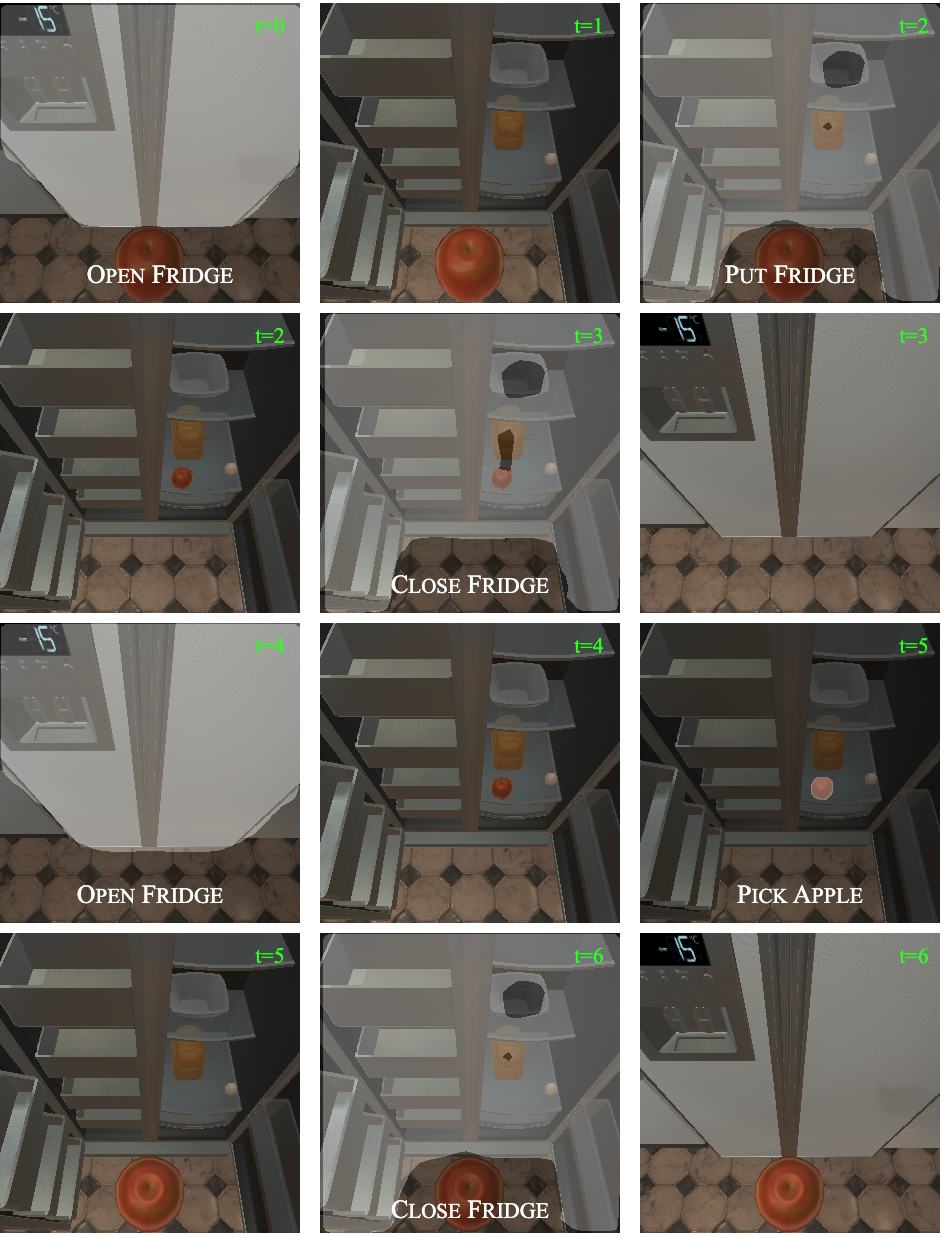}
    \end{subfigure}
    \caption{\textbf{Cool Object} interaction policy agent completes a subgoal task “Chill the apple in the refrigerator.” in an unseen environment.}
    \label{fig:cool}
\end{figure*}

\begin{figure*}[t!]
    \centering
    \begin{subfigure}{.8\linewidth}
        \centering
        \includegraphics[width=\linewidth]{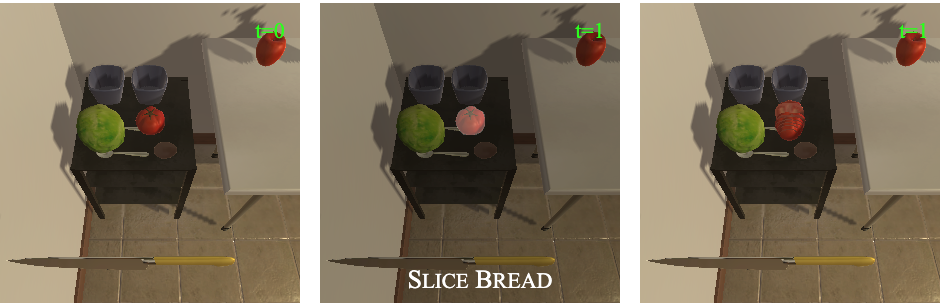}
    \end{subfigure}
    \caption{\textbf{Slice Object} interaction policy agent completes a subgoal task “Make slices of the bread.” in an unseen environment.}
    \label{fig:slice}
\end{figure*}

\begin{figure*}[t!]
    \centering
    \begin{subfigure}{.8\linewidth}
        \centering
        \includegraphics[width=\linewidth]{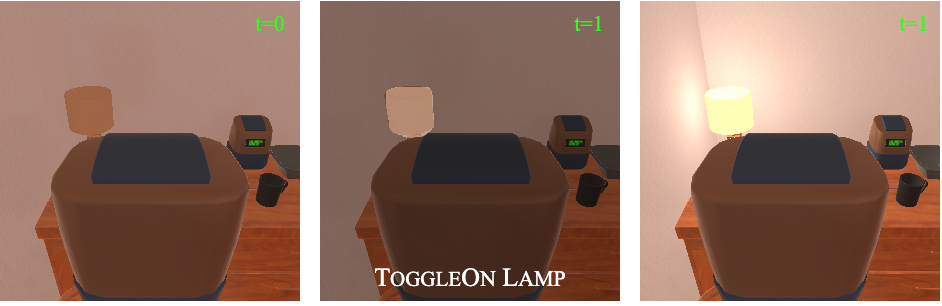}
    \end{subfigure}
    \caption{\textbf{Toggle Object} interaction policy agent completes a subgoal task “Turn on the lamp.” in an unseen environment.}
    \label{fig:toggle}
\end{figure*}

Table \ref{tab:subgoal} presents the performance of individual subgoal policies on corresponding subgoals.
Our hierarchical agent outperforms the flat policy by 25\% and 15\% average on the unseen and seen validation set respectively.
The modular nature of our interaction policies allows them to learn specific short-horizon subgoal tasks as illustrated in Fig. \ref{fig:pickup}-\ref{fig:toggle}, which reduces the complexity of the sequence of interaction actions and consequently makes the learning faster and easier.

\subsection{Task Type Evaluation}
\label{sec:add_expt}

There are seven high-level task categories in the ALFRED benchmark \cite{shridhar2020alfred}. Table \ref{tab:task_type} presents the quantitative comparison of our approach with a flat policy, a seq2seq model \cite{shridhar2020alfred}, and prior hierarchical baselines \cite{zhang2021hierarchical,blukis2021persistent} for each task type.
HLSM outperforms our agent on short-horizon tasks like `Pick \& Place' and `Examine.' 
For HLSM, these short-horizon tasks contribute towards most of the successful trajectories. 
The two most complex and longest-horizon task categories in the benchmark are `Stack \& Place' and `Pick Two \& Place.'
On these task types, \method gives an unseen success rate of 12.8\% and 23.4\%, respectively, compared to HLSM's 4.4\% and 18.0\%.
Overall, \method outperforms all the baselines in terms of average seen and unseen task success rates.

\subsection{Design Component Ablations}
\label{sec:supp_ablation}

We further provide ablation studies of design components of \method in Table \ref{tab:supp_ablation}.

\subsubsection{Data augmentation.}
We start by ablating the data augmentations described in Sec. `Implementation Details,' which lowers the performance across the board, highlighting its usefulness in training a smarter agent by reducing imitation learning sample complexity.
This results in a performance drop of 12.6\% and 25.0\% in the seen and unseen task success rates respectively.

\subsubsection{Factorised interaction policies.}
\rev{Then, inside the interaction policies, we replaced the decoupled pipeline \cite{singh2021factorizing} with a unified pipeline for both action sequence prediction and interaction mask generation.
Due to shared processing for mask and action modules, the ablated model's performance drops substantially, showing its inability to effectively utilize the input information.}

\subsubsection{Object-centric mask prediction.}
\rev{Finally, we train the interaction policies without the use of object-class-based mask prediction. as described in Sec. `Interaction Policy.'
Similar to \cite{shridhar2020alfred}, we directly upsample the joint embedding using deconvolution layers to generate the interaction mask.
We see a significant decline in performance on both seen and unseen folds due to poor mask generation ability, emphasising the need for object information for accurate mask formation.}

\begin{table}[t!]
    \centering
    \setlength{\tabcolsep}{2pt}
    \resizebox{\columnwidth}{!}{
        \begin{tabular}{@{}lcrccrc}
            \toprule
            \multirow{2}{*}{Input Ablations} && \multicolumn{2}{c}{\bf Validation-Seen}  & & \multicolumn{2}{c}{\bf Validation-Unseen}  \\
            \cmidrule{3-4} \cmidrule{6-7}
            && Task & GC && Task & GC  \\
            \cmidrule{1-7} 
            
            All Inputs (\method) &&	34.39 &	41.96 &&	20.08 &	38.43 \\
            {~~~-Egocentric View Only} &&	22.82 &	27.79 &	&12.51 &	23.87 \\
            {~~~-No Vision} &&	0.46 &	5.29 &&	0.25 &	8.31 \\
            {~~~-No Language} &&	2.74 &	8.21 &&	1.23 &	7.83 \\
            {~~~-Goal-Only} &&	6.39 &	12.7 &&	2.1	 &10.11 \\
            {~~~-Navigation Instruction only} &&	9.13 &	18.25 &&	3.07 &	15.05 \\
            \bottomrule
        \end{tabular}
    }
    \vspace{-0.5em}
    \caption{
        \textbf{Input Ablations.}
        We ablate different inputs to our network and report the task success rates and GC success rates.
        \revv{
            `Egocentric View Only' denotes training without surrounding perception.
            `No Vision' and `No Language' denote training without visual and textual input.
            `Goal-Only' denotes training without step-by-step instructions.
            `Navigation Instruction only' denotes that the master policy receives only navigational instructions.
            `w/o Loop Escape Module' denotes \method without the loop escape module.
        }
    }
    \label{tab:input_ablation}
\end{table}

\subsection{Input Ablation} 
\label{sec:input_ablation}
To further investigate our agent's vision and language bias, we provide ablation studies on inputs in Table \ref{tab:input_ablation}. 

\subsubsection{Egocentric View only.} 
This ablation specifies the condition under which the agent is deprived of its surrounding views (Sec. `Surrounding Perception`) while navigating without affecting other modules.
The agent's performance drops drastically, which demonstrates the significance of peripheral perception in novel environments.

\subsubsection{No Language.} 
We discover that when the agent receives only visual inputs without the language instructions, it is still capable of doing some tasks by memorising common sequences and target classes in the ‘seen’ fold. but does not adapt to the unseen environment.

\subsubsection{No Vision.}
This ablation demonstrates that the agent can still perform some subgoals, mostly navigation, by following the instructions to move along a few short trajectories, but is not able to complete goal tasks because it involves mask generation (object selection), which is entirely dependent on visual observation. 

\subsubsection{Goal-Only.} 
We define the setting where both the navigation and interaction policies receive only the goal task information. It gives a general idea of the task but doesn't contain any information about the tedious action combinations that the agent needs to perform to get the task done. This leads to a significant depletion in the task success rate.

\subsubsection{Navigation Instruction only.} 
This refers to the setting when the master policy only receives the navigation instruction corresponding to the current navigational subgoal. The agent can somewhat navigate but does not have the object information that acts as the stopping criterion which hinders the agent’s ability to perform interaction at the correct place in the environment. This demonstrates the benefit of our object-centric navigational approach.

\begin{table}[t!]
    \centering
    \setlength{\tabcolsep}{2pt}
    \renewcommand{\arraystretch}{1.1}
    \resizebox{\columnwidth}{!}{
        \begin{tabular}{@{}l|c|crccrc@{}}
            \toprule
            
            \multirow{2}{*}{\rev{Input Ablations}} & \multirow{1}{*}{\rev{PCC}} && \multicolumn{2}{c}{\bf \rev{Validation-Seen}}  & & \multicolumn{2}{c}{\bf \rev{Validation-Unseen}} \\
            
            \cline{4-5} \cline{7-8}
            \rule{0pt}{10pt} & Accuracy && \multicolumn{1}{c}{\rev{Task}} & \multicolumn{1}{c}{\rev{GC}} & & \multicolumn{1}{c}{\rev{Task}} & \multicolumn{1}{c}{\rev{GC}} \\
            \hline
            
            \rule{0pt}{10pt} \rev{\method} & $\B{98.4}$ && $\B{34.39}$  & $\B{41.96}$  && $\B{20.08}$  & $\B{38.42}$ \\ 
            
            $~~~-~$\rev{PCC w/ vision}& \rev{$96.5$} && \rev{$33.52$} & \rev{$40.56$} && \rev{$19.49$} & \rev{$37.48$} \\
            
            $~~~-~$\rev{MP w/o vision}& $-$  && ${1.21}$  & ${9.65}$  && ${0.85}$  & ${11.90}$ \\
            
            \bottomrule
        \end{tabular}
    }
    \caption{
        \rev{
        \textbf{Vision Ablations}
        ``PCC w/ vision'' denotes the \method agent using additional visual input for PCC.
        The second column depicts the performance of PCC for predicting the subgoal sequence evaluated on the validation set.
        ``MP w/o vision'' denotes the \method agent without visual input for master policy.
        The highest values per fold and task are shown in \B{blue}.
        }
    }
    \vspace{-1em}
    \label{tab:pcc_vision}
    \label{tab:mp_vision}
    
\end{table}

\subsubsection{Visual Input for Policy Composition Controller.}
The impact of visual features on PCC is presented in Table \ref{tab:pcc_vision}.
The agent does not gain from accessing visual information, as evidenced by the drop in performance for PCC w/ vision.
We believe that this happens due to the `causal misidentification' phenomenon \cite{de2019causal}; when receiving more information (\textit{e.g.}, RGB observation), the agent could learn to exploit irrelevant information (\textit{e.g.}, a chair in the room) to a target task (\textit{e.g.}, Heat a potato) rather than necessary information (\textit{e.g.}, microwave and potato), leading to a performance drop.

\subsubsection{Visual Input for Master Policy.}
\rev{The impact of visual features on the master policy is presented in Table \ref{tab:mp_vision}.
`MP w/o vision' demonstrates that the agent can still complete a few goal conditions, mostly the first navigation instruction, by following the instructions to move along a few short trajectories, but is not able to complete goal tasks because it involves interaction with objects in the environment (\ie, localising objects), which is entirely dependent on visual observation.}

\subsection{Ablation study for Loop Escape Module}
\label{supp:subsec:lem_ablation}

\begin{table}[h!]
    \centering
    \setlength{\tabcolsep}{4.5pt}
    \resizebox{\columnwidth}{!}{
        \begin{tabular}{@{}lccccccc}
            \toprule
             & \multirow{2}{*}{LEM} && \multicolumn{2}{c}{\bf Validation-Seen}  & & \multicolumn{2}{c}{\bf Validation-Unseen}  \\
            \cmidrule{4-5} \cmidrule{7-8}
            &&& SR & PLWSR && SR & PLWSR  \\
            \midrule 
            \multirow{2}{*}{\method} & \cmark && 34.39 & 23.04 && 20.08 &	10.84 \\
            
             & \xmark && 34.01 & 16.81 && 20.01 & 7.91 \\
            \bottomrule
        \end{tabular}
    }
    \vspace{-0.5em}
    \caption{
        \textbf{Effect of loop escape module (LEM).} We observe LEM contributes to the efficiency of the embodied agent.
    }
    \label{tab:ablation_lem}
\end{table}

We provide the ablation for the loop escape module (LEM) in Table \ref{tab:ablation_lem}. We observe that LEM provides marginal gains in task completion (SR) but has a major effect in terms of efficiency (PLWSR). It provides a 2.93\% improvement in unseen environments and a 6.23\% gain in seen environments.
LEM contributes to efficiency by early detection of loops in the trajectory of the agent.

\subsection{Sensitivity Analysis of \textit{W} in Loop Escape}
\label{supp:subsec:W_ana}
\rev{
The impact of the value of $W$ for the loop escape module (LEM) is examined in Fig. \ref{fig:subgaol_training} by conducting a grid search.
We observe that our agent achieves the best path length weighted success rate with $W = 10$.
Note that the performance does not drastically change by altering values of $W$, which indicates that the agent's performance is not very sensitive to this hyperparameter.
}

\subsection{Qualitative Analysis}
\label{subsec:qual_eg}

\begin{figure*}[t!]
    \centering
    \includegraphics[width=\linewidth]{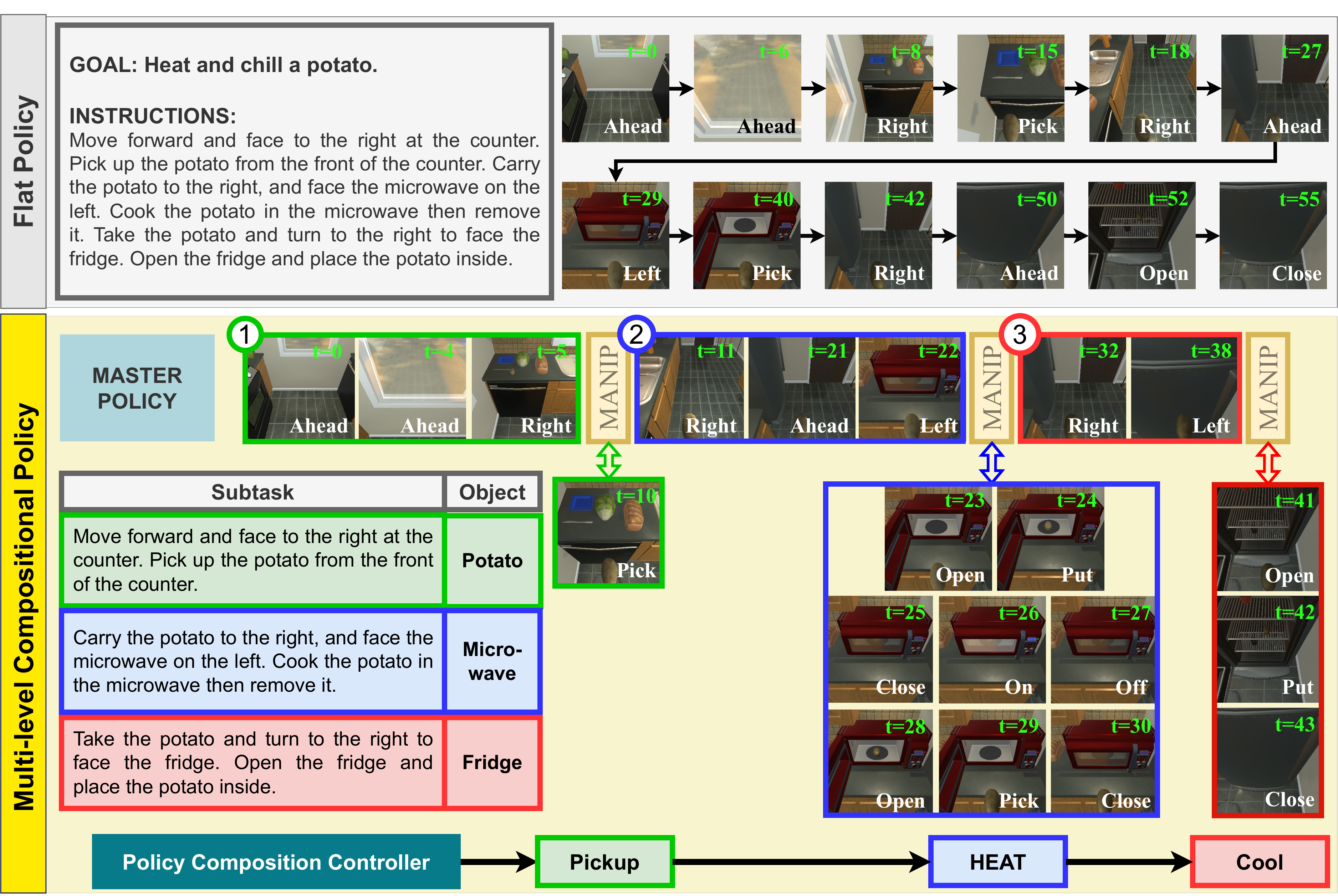}
    \caption{
        \textbf{Qualitative example for the complex task `Heat and chill a potato.'} 
    }
    \label{fig:interpret}
\end{figure*}

Fig. \ref{fig:interpret} shows an example inference of the hierarchical policy and the flat policy for a task, ``\textit{Heat and chill a potato.}''
The hierarchical agent decomposes instructions into subtasks, each of which includes a navigation and an interaction subgoal.
The PCC at the bottom outputs a sequence of subgoals used for selecting the appropriate interaction policy.
The object column demonstrates the output of the object encoding module.
The master policy uses the subtask instruction, object encoding, and visual observation to infer a navigation action at every time step as shown at the top.
It predicts the \texttt{<MANIPULATE>} token to shift the agent's control to the interaction policies along the way.
On completion of the interaction task, the control is shifted back to the navigation policy.
The hidden state of the navigation policy keeps a continuous encoding of navigation history without any interference of the interaction policy whereas each interaction policy uses an independent hidden state.
The flat policy shown at the top, on the other hand, shares the same parameters for learning both navigation and interaction subtasks.
This disentanglement results in simpler and faster learning by enabling parallel computation for various policies.

\begin{figure}[t!]
    \centering
    \includegraphics[width=.9\linewidth]{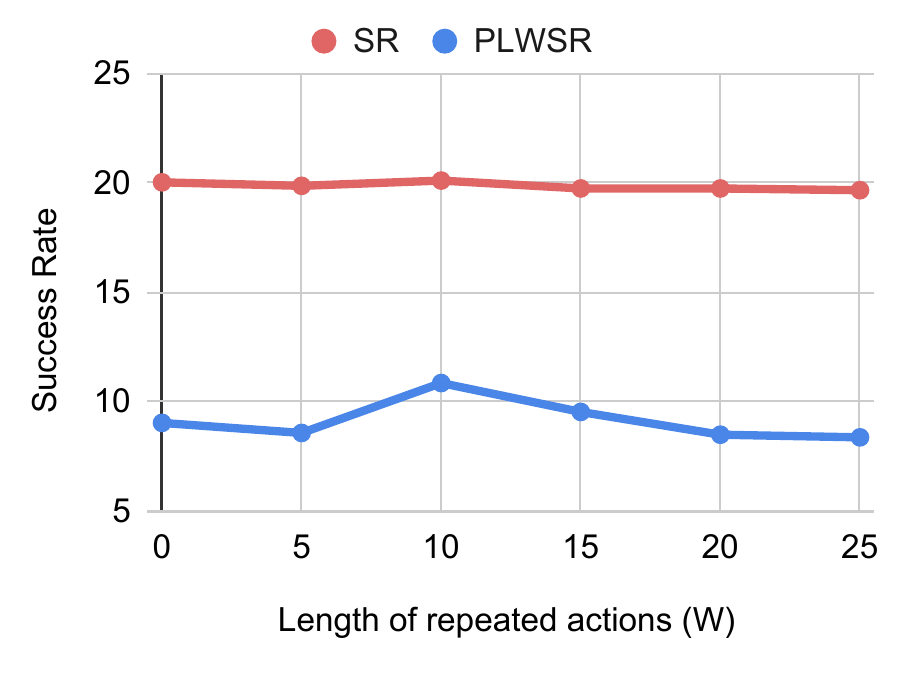}
    \vspace{-1.5em}
    \caption{
        \textbf{Sensitivity curve of $W$.}
        This figure shows the sensitivity analysis curve for the hyperparameter \textit{W} in the loop escape module described in Sec. `Loop Escape Module.'
    }
    \label{fig:W_ana}
    \vspace{-1.5em}
\end{figure}

\section{Extended Related Works}
\label{sec:ext_rw}

We provide a detailed technical comparison with the prior arts involving hierarchical frameworks \cite{zhang2021hierarchical,blukis2021persistent}.

\textbf{Comparison with HLSM \cite{blukis2021persistent}}
Here we compare the technical aspects of \method and HLSM.
Both works use action hierarchy to decompose trajectory into high-level subgoal actions and subgoals into low-level action sequences.
There are differences between the subgoals anticipated by \method and HLSM.
HLSM only employs interaction subgoals and considers navigation to be part of the same subgoal to reach the place of interaction, whereas \method includes an extra navigation subgoal `GoTo (object)' in which the agent navigates the environment to discover the object of interest. \method also contains the compound subgoals for Clean, Cool, and Heat, whereas HLSM would forecast the complete sequence of primary subgoals like Put, Pick, Toggle, etc.

In terms of model architecture, HLSM defines the task layout by using a hierarchical model with high-level and low-level controllers. The navigation and interaction components of a particular subgoal are collaboratively worked on by the low-level controller. \method, on the other hand, adopts a three-level hierarchy with independent modules for each level, namely, PCC, which provides the high-level structure of the trajectory, MP, which is responsible for navigation, and IP, which is composed of modules specialised for each interaction task.

We believe that HLSM and \method are complementary as HLSM uses depth supervision to estimate the 3D layout of the environment for detailed visual understanding while our \method exploits high-level language instructions for action planning. The depth supervision could be complementary to low-level instructions because the supervision of HLSM provides rich visual information by depth, which contains visual details, while the supervision of \method provides abstract language information for high-level guidance for actions. We argue that it is nontrivial to claim which supervision is less or more beneficial.
Using the object encoding module, \method also adds a navigational subgoal monitor that checks the availability of objects in the scene before executing an interaction policy.

\textbf{Comparison with HiTUT \cite{zhang2021hierarchical}} 
\rev{We also compare the technical aspects of \method and HiTUT. HiTUT also employs action hierarchy to decompose trajectory into high-level subgoal actions and subgoals into low-level action sequences like \method and HLSM. Even though HiTUT divides the task into a hierarchy of high-level subgoals and low-level actions, they utilize a unified transformer model with multiple heads to predict the subgoals and actions. In contrast, \method consists of independent modules for subgoal planning, navigation, and varying interaction tasks. The modular design allows for separate processing via specialised modules, making the learning process simpler due to short-term reasoning and enables the easy inclusion of additional interaction subgoals.}

\section{Discussion}
\label{sup:sec:discussion}

\subsection{What to learn \textit{vs.} what to bake in as inductive bias?}
\label{subsec:inductive_bias}
We agree that our method is a more or less ‘classical’ pipelined approach and has carefully designed components as a set of inductive biases. We attribute our design to (1) a lack of sufficient training data and (2) the requirement of heavy computational cost. (1) Given our framework is behavior cloning (BC) \cite{bain1995framework}, though the benchmark we used is the largest in the literature, the rigid nature of the BC approach (\textit{i.e.}, any deviation from expert trajectory over time is not allowed to learn a behavior) requires a vast amount of training examples (\textit{i.e.}, maybe not sufficient to be the largest dataset in literature) to learn a \textit{satisfactory} model. (2) To overcome the rigidity of learning a BC model, there are many approaches proposed in the literature \cite{ho2016generative,fu2017learning}. But they are mostly computationally expensive \cite{kostrikov2018discriminator}. We aim for the niche between both problems by designing well-crafted components for better encoding of the given dataset even in the rigid BC framework.

Our inductive bias could be useful for this task not just for this benchmark by the analogy of convolutional neural networks for images. By the universal approximation theorem, the 2D information, \textit{e.g.,} images, could be encoded by a multi-layer perceptron (MLP) with a sufficient amount of data. Given that it is difficult to fathom the sufficient amount of data to train the MLP, we have a very successful design choice for encoding 2D information; the patch-wise encoding and convolutional operations for its processing. Now, with 1M images, \textit{e.g.}, the ImageNet-1K, we can have a decently performing image classifier. We hope that our designed components could be useful to inspire future research in designing a data-driven model to address the Embodied AI task successfully. In sum, we pose our work as a stepping stone for designing and learning a successful data-driven model in the future.

\subsection{Hardware/training/computational cost of separate networks?}
\label{subsec:costs}

\subsubsection{Hardware Cost.}
\rev{
Our model would increase the memory cost linearly by the number of subgoals (0.23GB per subgoal) because models for each policy should be loaded in the memory proportional to the number of subgoals. Considering the large-sized memory in modern computer systems (\textit{e.g.}, usually larger than 8GB for PC and 6GB for high-end mobile phones), the additional memory cost may not be considered large, but it is certainly a cost.}

\rev{The training cost for the policy may marginally or may not increase because we divide a task into multiple subgoals and learning the specialized model for each subgoal takes fewer epochs as the training is \textit{easier} (\textit{i.e.}, sharing similar semantics) than that for the \textit{flat model}.
In Fig. \ref{fig:HvsF1}, we show that learning the specialized networks requires fewer training epochs compared to learning a unified network.
Therefore, even though there are multiple subgoals to be learned, the total training cost would increase marginally (and even embarrassingly parallelizable) or even decrease thanks to the reduced training cost per each subgoal added to an overhead of predicting subgoals.}

\subsubsection{Computational Cost.}
\rev{
The computational cost may marginally increase or even decrease because specialized networks for subgoals could infer each subgoal efficiently by reducing unnecessary exploration.
Fig. \ref{fig:HvsF2} shows that the specialized networks require shorter mean episode lengths to accomplish tasks.
With the efficient subgoal achievement, the overall computational cost at inference would be similar or even decrease, compared to the flat-policy model.
}

\begin{figure*}[t!]
    \centering
    \begin{subfigure}{.8\linewidth}
        \centering
        \includegraphics[width=\linewidth]{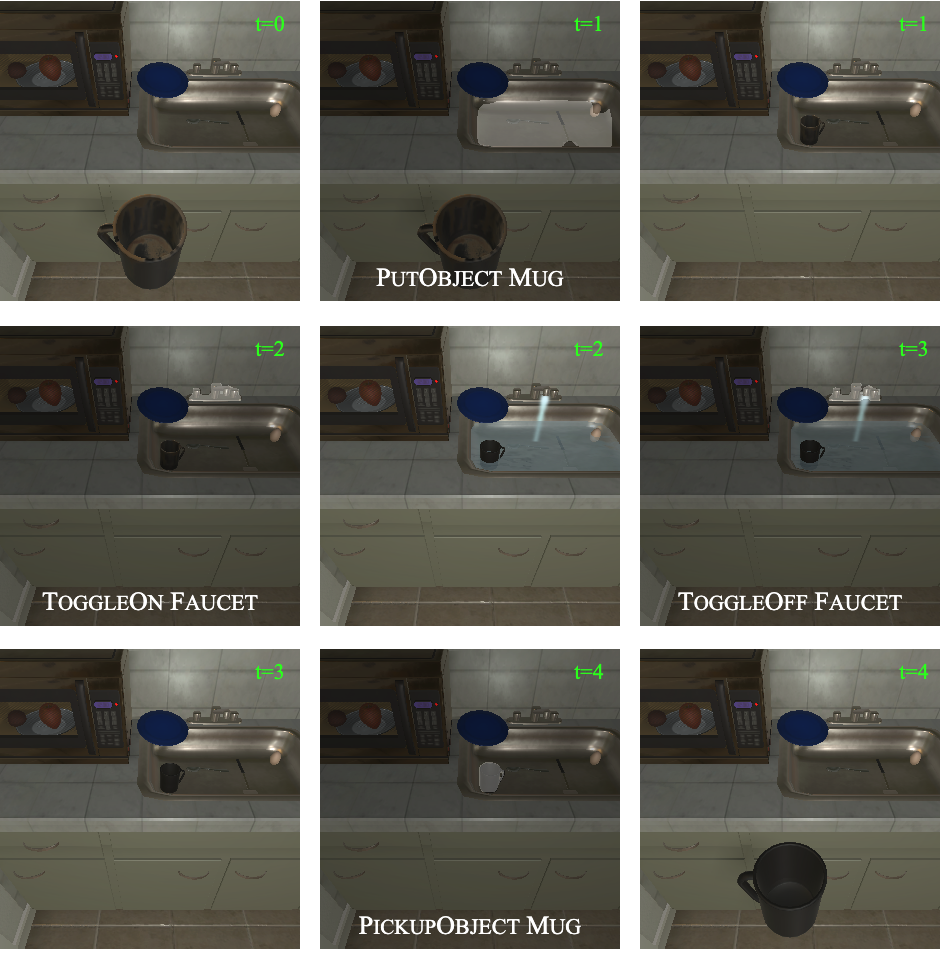}
    \end{subfigure}
    \caption{\textbf{Clean Object} interaction policy agent completes a subgoal task “Clean the mug in the sink.” in an unseen environment.}
    \label{fig:clean}
\end{figure*}

\end{document}